\documentclass[12pt]{format/IOP/iopart}
\usepackage{float}
\usepackage{color}
\usepackage{appendix}
\usepackage{booktabs}
\usepackage{graphicx}
\graphicspath{{../images/}}
\usepackage{doi}
\expandafter\let\csname equation*\endcsname\relax
\expandafter\let\csname endequation*\endcsname\relax
\usepackage{amsmath}
\usepackage{amsfonts}
\usepackage{amssymb}
\usepackage{subcaption}
\usepackage{adjustbox}
\usepackage{soul}
\graphicspath{{./images/}}
\usepackage[numbers,sort,compress]{natbib}
\usepackage{multirow}

\begin{document}
\def\allfiles{}
\ifx\allfiles\undefined
\input{format/packages-AIP}
\begin{document}
\fi

\title[SUTD-PRCM Dataset and NAS Approach for Complex Metasurface Design]{SUTD-PRCM Dataset and Neural Architecture Search Approach for Complex Metasurface Design}

\author{Tianning Zhang$^{1}$, Yee Sin Ang$^{1}$, Erping Li$^2$, Chun Yun Kee$^{1,*}$, L. K. Ang$^{1,*}$}

\address{$^1$Science, Mathematics and Technology, Singapore University of Technology and Design (SUTD), 8 Somapah Road, Singapore 487372}

\address{$^2$ College of Information Science and Electronic Engineering, Zhejiang University, Hangzhou 310027, China}

\address{$^*$Authors to whom any correspondence should be addressed}
\ead{chunyun\_kee@sutd.edu.sg, ricky\_ang@sutd.edu.sg}

\begin{abstract}
Metasurfaces have received a lot of attentions recently due to their versatile capability in manipulating electromagnetic wave. 
Advanced designs to satisfy multiple objectives with non-linear constraints have motivated researchers in using machine learning (ML) techniques like deep learning (DL) for accelerated design (forward and inverse) of metasurfaces.
For metasurfaces, it is difficult to make quantitative comparisons between different ML models without having a common and yet complex dataset used in many disciplines like image classification. 
Many studies were directed to a relatively constrained datasets that are limited to specified patterns or shapes in metasurfaces. 
In this paper, we present our SUTD polarized reflection of complex metasurfaces (SUTD-PRCM) dataset, which contains approximately 260,000 samples of complex metasurfaces created from electromagnetic simulation, and it has been used to benchmark our DL models.
The metasurface patterns are divided into different classes to facilitate different degree of complexity, which involves identifying and exploiting the relationship between the patterns and the electromagnetic responses that can be compared in using different DL models.
With the release of this SUTD-PRCM dataset, we hope that it will be useful for benchmarking 
existing or future DL models developed in the ML community.
We also propose a classification problem that is less encountered and apply neural architecture search (NAS) to have a preliminary understanding of potential modification to the neural architecture that will improve the prediction by DL models.
Our finding shows that convolution stacking is not the dominant element of the neural architecture anymore, which  implies  that  low-level  features  are  preferred  over the traditional deep  hierarchical  high-level features thus explains why deep convolutional neural network based models are not performing well in our dataset (SUTD-PRCM dataset).

\end{abstract}

\maketitle

\ifx\allfiles\undefined
\end{document}
\fi
\ifx\allfiles\undefined
\input{format/packages-Normal}
\begin{document}
\fi

\section{Introduction}

Due to the interaction between an electromagnetic (EM) wave and metasurfaces in some specific geometrical arrangements, metasurfaces can exhibit remarkable electromagnetic wave responses that have attracted great interests \cite{quevedo2019roadmap,li2020metamaterials,li2021metasurfaces,cui2017information}.
Metasurfaces have served as an important technology in many applications such as heat transforming \cite{li2021transforming},
cloaking \cite{yang2016full,qian2021perspective},
hologram \cite{liu2020work},
conversion \cite{liu2016fully},
absorption \cite{mitrofanov2018efficient,mitrofanov2020perfectly},
scattering reduction \cite{zhang2021hyperuniform},
polarization \cite{khan2017ultra,khorasaninejad2016metalenses,khorasaninejad2017metalenses},
transmission \cite{you2020broadband},
color \cite{cheng2015structural,proust2016all},
metalense \cite{khorasaninejad2016metalenses,khorasaninejad2017metalenses},
programmable metasurfaces \cite{bao2021programmable,zhang2020polarization,zhang2020optically},
and many others \cite{vasic2021refractive,
lin2020inverse,
niu2021dual,
al2021nature}.
These applications are made possible by the rapid advancement in micro- and even nano- fabrication technologies and computational modeling over the past decades. 
In the design of complex metasurfaces, machine learning (ML) methods like deep learning (DL) techniques has demonstrated unprecedented performance in providing rapid yet accurate prediction \cite{Omar2021deep}. 
Particularly, DL technique has been mainly applied for forward modeling and inverse design generation \cite{shi2020metasurface,mall2020fast,ding2022spatial,zhou2022metamaterials}. 
For forward modeling, instead of solving explicitly the governing Maxwell equations, DL models are capable of learning the complex non-linear mapping between input parameters to output EM response for a sufficiently large and high-quality dataset.
The resulting high fidelity surrogate model can readily replace the costly numerical solvers in the traditional design methodology based on evolutionary algorithms such as genetic algorithm (GA) \cite{jafar2018adaptive}, particle swarm optimization (PSO) \cite{zhang2017shaping}, and ant colonization optimization (ACO) \cite{zhu2019optimal}.
Compared to computationally expensive numerical solvers, efficient and accurate evaluation of DL surrogate model can lead to faster computational time and larger design search dimensions. 
In the inverse design, the generative model of desirable metasurfaces can be based on DL model with the desirable EM response as an input.
For a generative adversarial network (GAN) system, training of the generative model will involve a forward model which can be either numerical solvers or DL models. 
Similarly, replacing the forward model by a accurate and efficient DL model can lead to tremendous speed up in terms of training.
Nowadays, with metasurface designs typically represented as digital images, the most commonly adopted neural architectures in DL are deep convolutional neural network (DCNN) which is proven effective in various computer vision (CV) related problems.


All the advantages mentioned requires a good and if possible common dataset of metasurfaces to benchmark different DL models used in the community.
The involvement of DL typically started from the exploration with the most basic neural architecture, a fully connected network (FCN) \cite{malkiel2018plasmonic,peurifoy2018nanophotonic,an2019deep} for supervised learning.
With this approach, the electromagnetic scattering behaviour of an alternating dielectric thin films parameterized on thicknesses and dielectric constants of the films were successfully predicted \cite{liu2018training}.
In dealing with the instability and inconsistency problem, a bidirectional encoder-decoder model (Tandem) is proposed \cite{liu2018training}.
The perception of treating metasurfaces as images has led to seamless introduction of convolutional neural network (CNN) into metasurface design allowing 2D image as input.
A recent paper \cite{asano2018optimization} studied this problem when the output is only a scalar parameter.
ML algorithm on densely sampled spectral output such as reflection and transmission were also tested \cite{jiang2019free,jiang2019global}.
For inverse design, deep generative models are employed for generating new meta-atom designs to achieve the desired EM response.
Various groups \cite{jiang2019free,goodfellow2014generative} have used GAN system to quantify a differential mapping from desired EM response to the discrete 2D pattern.
In a recent paper \cite{haninverse}, contrast-vector is used to emphasize on the location of spectral peak in order to improve the performance of inverse design.
Another paper \cite{sajedian2019finding} further enhances the expression capability of the DL model by appending to CNN a recurrent neural network (RNN) which is more often seen in sequence modeling.
  
However, we observe that majority of the DL related works in metasurface design are restricted to canonical shapes or connected polygon which belong to a relatively simple and limited design dimension.
The findings reported in using such limited dataset are also qualitative at best in the comparison of different neural architectures. 
It is less intuitive to make meaningful comparison across different models without a common and more complex dataset. 
Easy access to such standard dataset will allow a more quantitative and fair comparison between different neural architectures and training strategies in the research community.
Inspired by how the standard datasets in CV community has advanced the state-of-the-art in their field, we are ready to share our dataset (SUTD-PRCM), which was first generated in a previous work and tested on different DL models \cite{zhang2021deep}. 
This dataset is essentially a collection of numerical simulated results of EM wave reflection of randomly created metasurfaces. 
Each sample in the dataset consists of an input metasurface of 16 x 16 binary image, and its associated output EM reflection as a function of frequency from 2 to 10 GHz. 
The randomly generated samples are sufficiently complex that are suitable for forward prediction and inverse design in testing different DL models in a simple GPU.
In our recent work \cite{zhang2021deep}, we have demonstrated that in using this SUTD-PRCM dataset tested with some existing DCNN based neural architectures that might not be the most optimal neural architecture yet. 
Thus this dataset is published here to share with the community for further testing.

For the first part of this paper, we introduce this SUTD-PRCM dataset in more details. We then present an automated approach to improve the architecture of DL models for better performance.
For demonstration purpose, an application to a classification problem based on this dataset is considered.
Note that the approach is general and it is likewise applicable to a regression problem \cite{zhang2021deep}.
Thus the objective of this paper is two folds. 
Firstly, we would like to share this physics based dataset (obtained by EM solvers) of metasurfaces with the community to explore future improvement in neural architecture for forward modeling and inverse design of such complex metasurface.
Relevant resources including the dataset and code for retrieving the data  are shared on Github.
Secondly, we will elaborate on the implementation of network architecture search (NAS) using a classification problem based on this dataset. 
The paper is organized as follows. 
The second section presents the details of the SUTD-PRCM dataset. 
The third section introduces several formulations of machine learning problems based on this dataset and relevant treatments needed. 
The fourth section considers one of the formulated ML problem in third section and NAS is applied to achieve better performance. 
Finally, we conclude the paper with a summary and possible future works.

\ifx\allfiles\undefined
\newpage
\bibliographystyle{unsrt}

\bibliography{
@article{cheng2015structural,
  title={Structural color printing based on plasmonic metasurfaces of perfect light absorption},
  author={Cheng, Fei and Gao, Jie and Luk, Ting S and Yang, Xiaodong},
  journal={Scientific reports},
  volume={5},
  pages={11045},
  year={2015},
  publisher={Nature Publishing Group}
}

@article{vasic2021refractive,
  title={Refractive index sensing with hollow metal--insulator--metal metasurfaces},
  author={Vasi{\'c}, Borislav and Isi{\'c}, Goran},
  journal={Journal of Physics D: Applied Physics},
  volume={54},
  number={28},
  pages={285106},
  year={2021},
  publisher={IOP Publishing}
}
@article{lin2020inverse,
  title={Inverse design of plasmonic metasurfaces by convolutional neural network},
  author={Lin, Ronghui and Zhai, Yanfen and Xiong, Chenxin and Li, Xiaohang},
  journal={Optics Letters},
  volume={45},
  number={6},
  pages={1362--1365},
  year={2020},
  publisher={Optical Society of America}
}
@article{niu2021dual,
  title={Dual-band and dual-polarized programmable metasurface unit with independent channels},
  author={Niu, Ling Yun and Zhang, Hao Chi and He, Pei Hang and Tang, Min and Wang, Meini and Bai, Guo Dong and Mao, Junfa and Cui, Tie Jun and others},
  journal={Journal of Physics D: Applied Physics},
  volume={54},
  number={14},
  pages={145109},
  year={2021},
  publisher={IOP Publishing}
}
@article{al2021nature,
  title={Nature-inspired orbital angular momentum beam generator using aperiodic metasurface},
  author={Al-Nuaimi, Mustafa K Taher and Hong, Wei and Whittow, William G},
  journal={Journal of Physics D: Applied Physics},
  volume={54},
  number={27},
  pages={275106},
  year={2021},
  publisher={IOP Publishing}
}

@article{gupta1998peak,
  title={Peak decomposition using Pearson type VII function},
  author={Gupta, SK},
  journal={Journal of applied crystallography},
  volume={31},
  number={3},
  pages={474--476},
  year={1998},
  publisher={International Union of Crystallography}
}
@article{ding2022spatial,
  title={Spatial phase retrieval of vortex beam using convolutional neural network},
  author={Ding, Ge and Xiong, Wenjie and Wang, Peipei and Huang, Zebin and He, Yanliang and Liu, Junmin and Li, Ying and Fan, Dianyuan and Chen, Shuqing},
  journal={Journal of Optics},
  volume={24},
  number={2},
  pages={025701},
  year={2022},
  publisher={IOP Publishing}
}

@inproceedings{zhou2022metamaterials,
  title={Metamaterials Design Method based on Deep learning Database},
  author={Zhou, Xiaoshu and Xiao, Qide and Wang, Han},
  booktitle={Journal of Physics: Conference Series},
  volume={2185},
  number={1},
  pages={012023},
  year={2022},
  organization={IOP Publishing}
}
@article{mall2020fast,
  title={Fast design of plasmonic metasurfaces enabled by deep learning},
  author={Mall, Abhishek and Patil, Abhijeet and Tamboli, Dipesh and Sethi, Amit and Kumar, Anshuman},
  journal={Journal of Physics D: Applied Physics},
  volume={53},
  number={49},
  pages={49LT01},
  year={2020},
  publisher={IOP Publishing}
}
@article{shi2020metasurface,
  title={Metasurface inverse design using machine learning approaches},
  author={Shi, Xin and Qiu, Tianshuo and Wang, Jiafu and Zhao, Xueqing and Qu, Shaobo},
  journal={Journal of Physics D: Applied Physics},
  volume={53},
  number={27},
  pages={275105},
  year={2020},
  publisher={IOP Publishing}
}

@article{mazzia2021efficient,
  title={Efficient-capsnet: Capsule network with self-attention routing},
  author={Mazzia, Vittorio and Salvetti, Francesco and Chiaberge, Marcello},
  journal={Scientific Reports},
  volume={11},
  number={1},
  pages={1--13},
  year={2021},
  publisher={Nature Publishing Group}
}

@article{bruna2013invariant,
  title={Invariant scattering convolution networks},
  author={Bruna, Joan and Mallat, St{\'e}phane},
  journal={IEEE transactions on pattern analysis and machine intelligence},
  volume={35},
  number={8},
  pages={1872--1886},
  year={2013},
  publisher={IEEE}
}
@article{sak2014long,
  title={Long short-term memory recurrent neural network architectures for large scale acoustic modeling},
  author={Sak, Hasim and Senior, Andrew W and Beaufays, Fran{\c{c}}oise},
  year={2014}
}
@article{lin2020inverse,
  title={Inverse design of plasmonic metasurfaces by convolutional neural network},
  author={Lin, Ronghui and Zhai, Yanfen and Xiong, Chenxin and Li, Xiaohang},
  journal={Optics letters},
  volume={45},
  number={6},
  pages={1362--1365},
  year={2020},
  publisher={Optical Society of America}
}
@article{bronstein2021geometric,
  title={Geometric deep learning: Grids, groups, graphs, geodesics, and gauges},
  author={Bronstein, Michael M and Bruna, Joan and Cohen, Taco and Veli{\v{c}}kovi{\'c}, Petar},
  journal={arXiv preprint arXiv:2104.13478},
  year={2021}
}

@inproceedings{cohen2016group,
  title={Group equivariant convolutional networks},
  author={Cohen, Taco and Welling, Max},
  booktitle={International conference on machine learning},
  pages={2990--2999},
  year={2016},
  organization={PMLR}
}
@article{vaswani2017attention,
  title={Attention is all you need},
  author={Vaswani, Ashish and Shazeer, Noam and Parmar, Niki and Uszkoreit, Jakob and Jones, Llion and Gomez, Aidan N and Kaiser, Lukasz and Polosukhin, Illia},
  journal={arXiv preprint arXiv:1706.03762},
  year={2017}
}

@article{liu2018darts,
  title={DARTS: Differentiable Architecture Search},
  author={Liu, Hanxiao and Simonyan, Karen and Yang, Yiming},
  journal={arXiv preprint arXiv:1806.09055},
  year={2018}
}
@article{kandasamy2018neural,
  title={Neural architecture search with bayesian optimisation and optimal transport},
  author={Kandasamy, Kirthevasan and Neiswanger, Willie and Schneider, Jeff and Poczos, Barnabas and Xing, Eric},
  journal={arXiv preprint arXiv:1802.07191},
  year={2018}
}

@article{li2020geometry,
  title={Geometry-Aware Gradient Algorithms for Neural Architecture Search},
  author={Li, Liam and Khodak, Mikhail and Balcan, Maria-Florina and Talwalkar, Ameet},
  journal={arXiv preprint arXiv:2004.07802},
  year={2020}
}

@article{xu2019pc,
  title={PC-DARTS: Partial channel connections for memory-efficient architecture search},
  author={Xu, Yuhui and Xie, Lingxi and Zhang, Xiaopeng and Chen, Xin and Qi, Guo-Jun and Tian, Qi and Xiong, Hongkai},
  journal={arXiv preprint arXiv:1907.05737},
  year={2019}
}

@inproceedings{real2017large,
  title={Large-scale evolution of image classifiers},
  author={Real, Esteban and Moore, Sherry and Selle, Andrew and Saxena, Saurabh and Suematsu, Yutaka Leon and Tan, Jie and Le, Quoc V and Kurakin, Alexey},
  booktitle={International Conference on Machine Learning},
  pages={2902--2911},
  year={2017},
  organization={PMLR}
}
@article{zoph2016neural,
  title={Neural architecture search with reinforcement learning},
  author={Zoph, Barret and Le, Quoc V},
  journal={arXiv preprint arXiv:1611.01578},
  year={2016}
}
@article{qin2019nasnet,
  title={Nasnet: A neuron attention stage-by-stage net for single image deraining},
  author={Qin, Xu and Wang, Zhilin},
  journal={arXiv preprint arXiv:1912.03151},
  year={2019}
}
@article{elsken2019neural,
  title={Neural architecture search: A survey.},
  author={Elsken, Thomas and Metzen, Jan Hendrik and Hutter, Frank and others},
  journal={J. Mach. Learn. Res.},
  volume={20},
  number={55},
  pages={1--21},
  year={2019}
}
@article{hodge2021deep,
  title={Deep inverse design of reconfigurable metasurfaces for future communications},
  author={Hodge, John A and Mishra, Kumar Vijay and Zaghloul, Amir I},
  journal={arXiv preprint arXiv:2101.09131},
  year={2021}
}
@article{battaglia2018relational,
  title={Relational inductive biases, deep learning, and graph networks},
  author={Battaglia, Peter W and Hamrick, Jessica B and Bapst, Victor and Sanchez-Gonzalez, Alvaro and Zambaldi, Vinicius and Malinowski, Mateusz and Tacchetti, Andrea and Raposo, David and Santoro, Adam and Faulkner, Ryan and others},
  journal={arXiv preprint arXiv:1806.01261},
  year={2018}
}

@misc{PolygonGenerate,
  title = {Algorithm to generate random 2D polygon},
  howpublished = {\url{https://stackoverflow.com/questions/8997099/algorithm-to-generate-random-2d-polygon}},
}
@article{proust2016all,
  title={All-dielectric colored metasurfaces with silicon Mie resonators},
  author={Proust, Julien and Bedu, Frederic and Gallas, Bruno and Ozerov, Igor and Bonod, Nicolas},
  journal={ACS nano},
  volume={10},
  number={8},
  pages={7761--7767},
  year={2016},
  publisher={ACS Publications}
}
@article{quevedo2019roadmap,
  title={Roadmap on metasurfaces},
  author={Quevedo-Teruel, Oscar and Chen, Hongsheng and D{\'\i}az-Rubio, Ana and Gok, Gurkan and Grbic, Anthony and Minatti, Gabriele and Martini, Enrica and Maci, Stefano and Eleftheriades, George V and Chen, Michael and others},
  journal={Journal of Optics},
  volume={21},
  number={7},
  pages={073002},
  year={2019},
  publisher={IOP Publishing}
}

@article{li2021metasurfaces,
  title={Metasurfaces for bioelectronics and healthcare},
  author={Li, Zhipeng and Tian, Xi and Qiu, Cheng-Wei and Ho, John S},
  journal={Nature Electronics},
  pages={1--10},
  year={2021},
  publisher={Nature Publishing Group}
}
@article{li2021transforming,
  title={Transforming heat transfer with thermal metamaterials and devices},
  author={Li, Ying and Li, Wei and Han, Tiancheng and Zheng, Xu and Li, Jiaxin and Li, Baowen and Fan, Shanhui and Qiu, Cheng-Wei},
  journal={Nature Reviews Materials},
  volume={6},
  number={6},
  pages={488--507},
  year={2021},
  publisher={Nature Publishing Group}
}
@article{zhu2021phase,
  title={Phase-to-pattern inverse design paradigm for fast realization of functional metasurfaces via transfer learning},
  author={Zhu, Ruichao and Qiu, Tianshuo and Wang, Jiafu and Sui, Sai and Hao, Chenglong and Liu, Tonghao and Li, Yongfeng and Feng, Mingde and Zhang, Anxue and Qiu, Cheng-Wei and others},
  journal={Nature communications},
  volume={12},
  number={1},
  pages={1--10},
  year={2021},
  publisher={Nature Publishing Group}
}
@article{zhang2020optically,
  title={An optically driven digital metasurface for programming electromagnetic functions},
  author={Zhang, Xin Ge and Jiang, Wei Xiang and Jiang, Hao Lin and Wang, Qiang and Tian, Han Wei and Bai, Lin and Luo, Zhang Jie and Sun, Shang and Luo, Yu and Qiu, Cheng-Wei and others},
  journal={Nature Electronics},
  volume={3},
  number={3},
  pages={165--171},
  year={2020},
  publisher={Nature Publishing Group}
}
@article{zhang2020polarization,
  title={Polarization-Controlled Dual-Programmable Metasurfaces},
  author={Zhang, Xin Ge and Yu, Qian and Jiang, Wei Xiang and Sun, Ya Lun and Bai, Lin and Wang, Qiang and Qiu, Cheng-Wei and Cui, Tie Jun},
  journal={Advanced science},
  volume={7},
  number={11},
  pages={1903382},
  year={2020},
  publisher={Wiley Online Library}
}
@book{li2020metamaterials,
  title={Metamaterials and Negative Refraction},
  author={Li, Rujiang and Wang, Zuojia and Chen, Hongsheng},
  year={2020},
  publisher={Cambridge University Press}
}
@article{cui2017information,
  title={Information metamaterials and metasurfaces},
  author={Cui, Tie Jun and Liu, Shuo and Zhang, Lei},
  journal={Journal of Materials Chemistry C},
  volume={5},
  number={15},
  pages={3644--3668},
  year={2017},
  publisher={Royal Society of Chemistry}
}
@article{yang2016full,
  title={Full-polarization 3D metasurface cloak with preserved amplitude and phase},
  author={Yang, Yihao and Jing, Liqiao and Zheng, Bin and Hao, Ran and Yin, Wenyan and Li, Erping and Soukoulis, Costas M and Chen, Hongsheng},
  journal={Advanced Materials},
  volume={28},
  number={32},
  pages={6866--6871},
  year={2016},
  publisher={Wiley Online Library}
}

@article{qian2021perspective,
  title={A perspective on the next generation of invisibility cloaks—Intelligent cloaks},
  author={Qian, Chao and Chen, Hongsheng},
  journal={Applied Physics Letters},
  volume={118},
  number={18},
  pages={180501},
  year={2021},
  publisher={AIP Publishing LLC}
}
@inproceedings{liu2020work,
  title={Work in Progress: Intelligent Metasurface Holograms},
  author={Liu, Che and Ma, Qian and Li, Lianlin and Cui, Tie Jun},
  booktitle={Proceedings of the 1st ACM International Workshop on Nanoscale Computing, Communication, and Applications},
  pages={45--48},
  year={2020}
}

@article{liu2016fully,
  title={Fully controllable Pancharatnam-Berry metasurface array with high conversion efficiency and broad bandwidth},
  author={Liu, Chuanbao and Bai, Yang and Zhao, Qian and Yang, Yihao and Chen, Hongsheng and Zhou, Ji and Qiao, Lijie},
  journal={Scientific reports},
  volume={6},
  number={1},
  pages={1--7},
  year={2016},
  publisher={Nature Publishing Group}
}
@article{mitrofanov2018efficient,
  title={Efficient photoconductive terahertz detector with all-dielectric optical metasurface},
  author={Mitrofanov, Oleg and Siday, Thomas and Thompson, Robert J and Luk, Ting Shan and Brener, Igal and Reno, John L},
  journal={APL Photonics},
  volume={3},
  number={5},
  pages={051703},
  year={2018},
  publisher={AIP Publishing LLC}
}
@article{mitrofanov2020perfectly,
  title={Perfectly absorbing dielectric metasurfaces for photodetection},
  author={Mitrofanov, Oleg and Hale, Lucy L and Vabishchevich, Polina P and Luk, Ting Shan and Addamane, Sadhvikas J and Reno, John L and Brener, Igal},
  journal={APL Photonics},
  volume={5},
  number={10},
  pages={101304},
  year={2020},
  publisher={AIP Publishing LLC}
}
@article{zhang2021hyperuniform,
  title={Hyperuniform disordered distribution metasurface for scattering reduction},
  author={Zhang, Haoyang and Cheng, Qiao and Chu, Hongchen and Christogeorgos, Orestis and Wu, Wen and Hao, Yang},
  journal={Applied Physics Letters},
  volume={118},
  number={10},
  pages={101601},
  year={2021},
  publisher={AIP Publishing LLC}
}
@article{khan2017ultra,
  title={Ultra-wideband cross polarization conversion metasurface insensitive to incidence angle},
  author={Khan, Muhammad Ismail and Fraz, Qaisar and Tahir, Farooq A},
  journal={Journal of Applied Physics},
  volume={121},
  number={4},
  pages={045103},
  year={2017},
  publisher={AIP Publishing LLC}
}
@article{you2020broadband,
  title={Broadband terahertz transmissive quarter-wave metasurface},
  author={You, Xiaolong and Ako, Rajour T and Lee, Wendy SL and Bhaskaran, Madhu and Sriram, Sharath and Fumeaux, Christophe and Withayachumnankul, Withawat},
  journal={APL Photonics},
  volume={5},
  number={9},
  pages={096108},
  year={2020},
  publisher={AIP Publishing LLC}
}
@article{bao2021programmable,
  title={Programmable Reflection--Transmission Shared-Aperture Metasurface for Real-Time Control of Electromagnetic Waves in Full Space},
  author={Bao, Lei and Ma, Qian and Wu, Rui Yuan and Fu, Xiaojian and Wu, Junwei and Cui, Tie Jun},
  journal={Advanced Science},
  pages={2100149},
  year={2021},
  publisher={Wiley Online Library}
}
@article{khorasaninejad2016metalenses,
  title={Metalenses at visible wavelengths: Diffraction-limited focusing and subwavelength resolution imaging},
  author={Khorasaninejad, Mohammadreza and Chen, Wei Ting and Devlin, Robert C and Oh, Jaewon and Zhu, Alexander Y and Capasso, Federico},
  journal={Science},
  volume={352},
  number={6290},
  pages={1190--1194},
  year={2016},
  publisher={American Association for the Advancement of Science}
}
@article{khorasaninejad2017metalenses,
  title={Metalenses: Versatile multifunctional photonic components},
  author={Khorasaninejad, Mohammadreza and Capasso, Federico},
  journal={Science},
  volume={358},
  number={6367},
  year={2017},
  publisher={American Association for the Advancement of Science}
}

@article{yu2012broadband,
  title={A broadband, background-free quarter-wave plate based on plasmonic metasurfaces},
  author={Yu, Nanfang and Aieta, Francesco and Genevet, Patrice and Kats, Mikhail A and Gaburro, Zeno and Capasso, Federico},
  journal={Nano letters},
  volume={12},
  number={12},
  pages={6328--6333},
  year={2012},
  publisher={ACS Publications}
}
@article{lecun2015deep,
  title={Deep learning},
  author={LeCun, Yann and Bengio, Yoshua and Hinton, Geoffrey},
  journal={nature},
  volume={521},
  number={7553},
  pages={436--444},
  year={2015},
  publisher={Nature Publishing Group}
}

@article{carrasquilla2017machine,
  title={Machine learning phases of matter},
  author={Carrasquilla, Juan and Melko, Roger G},
  journal={Nature Physics},
  volume={13},
  number={5},
  pages={431--434},
  year={2017},
  publisher={Nature Publishing Group}
}

@article{baldi2014searching,
  title={Searching for exotic particles in high-energy physics with deep learning},
  author={Baldi, Pierre and Sadowski, Peter and Whiteson, Daniel},
  journal={Nature communications},
  volume={5},
  number={1},
  pages={1--9},
  year={2014},
  publisher={Nature Publishing Group}
}

@article{segler2018planning,
  title={Planning chemical syntheses with deep neural networks and symbolic AI},
  author={Segler, Marwin HS and Preuss, Mike and Waller, Mark P},
  journal={Nature},
  volume={555},
  number={7698},
  pages={604--610},
  year={2018},
  publisher={Nature Publishing Group}
}

@article{chen2016deep,
  title={Deep learning in label-free cell classification},
  author={Chen, Claire Lifan and Mahjoubfar, Ata and Tai, Li-Chia and Blaby, Ian K and Huang, Allen and Niazi, Kayvan Reza and Jalali, Bahram},
  journal={Scientific reports},
  volume={6},
  pages={21471},
  year={2016},
  publisher={Nature Publishing Group}
}

@article{zhou2017pdeep,
  title={pDeep: predicting MS/MS spectra of peptides with deep learning},
  author={Zhou, Xie-Xuan and Zeng, Wen-Feng and Chi, Hao and Luo, Chunjie and Liu, Chao and Zhan, Jianfeng and He, Si-Min and Zhang, Zhifei},
  journal={Analytical chemistry},
  volume={89},
  number={23},
  pages={12690--12697},
  year={2017},
  publisher={ACS Publications}
}

@inproceedings{goodfellow2014generative,
  title={Generative adversarial nets},
  author={Goodfellow, Ian and Pouget-Abadie, Jean and Mirza, Mehdi and Xu, Bing and Warde-Farley, David and Ozair, Sherjil and Courville, Aaron and Bengio, Yoshua},
  booktitle={Advances in neural information processing systems},
  pages={2672--2680},
  year={2014}
}
@article{gessulat2019prosit,
  title={Prosit: proteome-wide prediction of peptide tandem mass spectra by deep learning},
  author={Gessulat, Siegfried and Schmidt, Tobias and Zolg, Daniel Paul and Samaras, Patroklos and Schnatbaum, Karsten and Zerweck, Johannes and Knaute, Tobias and Rechenberger, Julia and Delanghe, Bernard and Huhmer, Andreas and others},
  journal={Nature methods},
  volume={16},
  number={6},
  pages={509--518},
  year={2019},
  publisher={Nature Publishing Group}
}

@article{guo2016broadband,
  title={Broadband polarizers based on graphene metasurfaces},
  author={Guo, Tianjing and Argyropoulos, Christos},
  journal={Optics letters},
  volume={41},
  number={23},
  pages={5592--5595},
  year={2016},
  publisher={Optical Society of America}
}
@article{devlin2017arbitrary,
  title={Arbitrary spin-to--orbital angular momentum conversion of light},
  author={Devlin, Robert C and Ambrosio, Antonio and Rubin, Noah A and Mueller, JP Balthasar and Capasso, Federico},
  journal={Science},
  volume={358},
  number={6365},
  pages={896--901},
  year={2017},
  publisher={American Association for the Advancement of Science}
}
@article{sui2015topology,
  title={Topology optimization design of a lightweight ultra-broadband wide-angle resistance frequency selective surface absorber},
  author={Sui, Sai and Ma, Hua and Wang, Jiafu and Pang, Yongqiang and Qu, Shaobo},
  journal={Journal of Physics D: Applied Physics},
  volume={48},
  number={21},
  pages={215101},
  year={2015},
  publisher={IOP Publishing}
}

@article{zhu2019optimal,
  title={Optimal high efficiency 3D plasmonic metasurface elements revealed by lazy ants},
  author={Zhu, Danny Z and Whiting, Eric B and Campbell, Sawyer D and Burckel, D Bruce and Werner, Douglas H},
  journal={ACS Photonics},
  volume={6},
  number={11},
  pages={2741--2748},
  year={2019},
  publisher={ACS Publications}
}
@article{jafar2018adaptive,
  title={Adaptive genetic algorithm for optical metasurfaces design},
  author={Jafar-Zanjani, Samad and Inampudi, Sandeep and Mosallaei, Hossein},
  journal={Scientific reports},
  volume={8},
  number={1},
  pages={1--16},
  year={2018},
  publisher={Nature Publishing Group}
}
@article{liu2018training,
  title={Training deep neural networks for the inverse design of nanophotonic structures},
  author={Liu, Dianjing and Tan, Yixuan and Khoram, Erfan and Yu, Zongfu},
  journal={ACS Photonics},
  volume={5},
  number={4},
  pages={1365--1369},
  year={2018},
  publisher={ACS Publications}
}

@article{malkiel2018plasmonic,
  title={Plasmonic nanostructure design and characterization via deep learning},
  author={Malkiel, Itzik and Mrejen, Michael and Nagler, Achiya and Arieli, Uri and Wolf, Lior and Suchowski, Haim},
  journal={Light: Science \& Applications},
  volume={7},
  number={1},
  pages={1--8},
  year={2018},
  publisher={Nature Publishing Group}
}

@inproceedings{allen2017metasurface,
  title={Metasurface engineering via evolutionary processes},
  author={Allen, Kenneth W and Dykes, Daniel JP and Reid, David R and Bean, Jeffrey A and Landgren, David W and Lee, R Todd and Denison, Douglas R},
  booktitle={2017 IEEE National Aerospace and Electronics Conference (NAECON)},
  pages={172--178},
  year={2017},
  organization={IEEE}
}
@inproceedings{he2016deep,
  title={Deep residual learning for image recognition},
  author={He, Kaiming and Zhang, Xiangyu and Ren, Shaoqing and Sun, Jian},
  booktitle={Proceedings of the IEEE conference on computer vision and pattern recognition},
  pages={770--778},
  year={2016}
}
@article{peurifoy2018nanophotonic,
  title={Nanophotonic particle simulation and inverse design using artificial neural networks},
  author={Peurifoy, John and Shen, Yichen and Jing, Li and Yang, Yi and Cano-Renteria, Fidel and DeLacy, Brendan G and Joannopoulos, John D and Tegmark, Max and Solja{\v{c}}i{\'c}, Marin},
  journal={Science advances},
  volume={4},
  number={6},
  pages={eaar4206},
  year={2018},
  publisher={American Association for the Advancement of Science}
}

@article{an2019deep,
  title={A deep learning approach for objective-driven all-dielectric metasurface design},
  author={An, Sensong and Fowler, Clayton and Zheng, Bowen and Shalaginov, Mikhail Y and Tang, Hong and Li, Hang and Zhou, Li and Ding, Jun and Agarwal, Anuradha Murthy and Rivero-Baleine, Clara and others},
  journal={ACS Photonics},
  volume={6},
  number={12},
  pages={3196--3207},
  year={2019},
  publisher={ACS Publications}
}

@article{an2020freeform,
  title={A freeform dielectric metasurface modeling approach based on deep neural networks},
  author={An, Sensong and Zheng, Bowen and Shalaginov, Mikhail Y and Tang, Hong and Li, Hang and Zhou, Li and Ding, Jun and Agarwal, Anuradha Murthy and Rivero-Baleine, Clara and Kang, Myungkoo and others},
  journal={arXiv preprint arXiv:2001.00121},
  year={2020}
}

@article{tahersima2019deep,
  title={Deep neural network inverse design of integrated photonic power splitters},
  author={Tahersima, Mohammad H and Kojima, Keisuke and Koike-Akino, Toshiaki and Jha, Devesh and Wang, Bingnan and Lin, Chungwei and Parsons, Kieran},
  journal={Scientific reports},
  volume={9},
  number={1},
  pages={1--9},
  year={2019},
  publisher={Nature Publishing Group}
}
@article{liu2018generative,
  title={Generative model for the inverse design of metasurfaces},
  author={Liu, Zhaocheng and Zhu, Dayu and Rodrigues, Sean P and Lee, Kyu-Tae and Cai, Wenshan},
  journal={Nano letters},
  volume={18},
  number={10},
  pages={6570--6576},
  year={2018},
  publisher={ACS Publications}
}
@article{haninverse,
  title={Inverse design of metasurface optical filters using deep neural network with high degrees of freedom},
  author={Han, Xiao and Fan, Ziyang and Liu, Zeyang and Li, Chao and Guo, L Jay},
  journal={InfoMat},
  publisher={Wiley Online Library}
}
@article{asano2018optimization,
  title={Optimization of photonic crystal nanocavities based on deep learning},
  author={Asano, Takashi and Noda, Susumu},
  journal={Optics express},
  volume={26},
  number={25},
  pages={32704--32717},
  year={2018},
  publisher={Optical Society of America}
}
@article{wang2004image,
  title={Image quality assessment: from error visibility to structural similarity},
  author={Wang, Zhou and Bovik, Alan C and Sheikh, Hamid R and Simoncelli, Eero P},
  journal={IEEE transactions on image processing},
  volume={13},
  number={4},
  pages={600--612},
  year={2004},
  publisher={IEEE}
}
@article{sajedian2019finding,
  title={Finding the optical properties of plasmonic structures by image processing using a combination of convolutional neural networks and recurrent neural networks},
  author={Sajedian, Iman and Kim, Jeonghyun and Rho, Junsuk},
  journal={Microsystems \& nanoengineering},
  volume={5},
  number={1},
  pages={1--8},
  year={2019},
  publisher={Nature Publishing Group}
}

@article{ma2018deep,
  title={Deep-learning-enabled on-demand design of chiral metamaterials},
  author={Ma, Wei and Cheng, Feng and Liu, Yongmin},
  journal={ACS nano},
  volume={12},
  number={6},
  pages={6326--6334},
  year={2018},
  publisher={ACS Publications}
}

@article{jiang2019free,
  title={Free-form diffractive metagrating design based on generative adversarial networks},
  author={Jiang, Jiaqi and Sell, David and Hoyer, Stephan and Hickey, Jason and Yang, Jianji and Fan, Jonathan A},
  journal={ACS nano},
  volume={13},
  number={8},
  pages={8872--8878},
  year={2019},
  publisher={ACS Publications}
}
@article{jiang2019global,
  title={Global optimization of dielectric metasurfaces using a physics-driven neural network},
  author={Jiang, Jiaqi and Fan, Jonathan A},
  journal={Nano letters},
  volume={19},
  number={8},
  pages={5366--5372},
  year={2019},
  publisher={ACS Publications}
}
@article{zhang2021deep,
author = {Zhang,Tianning  and Kee,Chun Yun  and Ang,Yee Sin  and Ang,L. K. },
title = {Deep learning-based design of broadband GHz complex and random metasurfaces},
journal = {APL Photonics},
volume = {6},
number = {10},
pages = {106101},
year = {2021},
doi = {10.1063/5.0061571}

}

@article{kasim2021building,
	author = {M F Kasim and D Watson-Parris and L Deaconu and S Oliver and P Hatfield and D H Froula and G Gregori and M Jarvis and S Khatiwala and J Korenaga and J Topp-Mugglestone and E Viezzer and S M Vinko},
	title = {Building high accuracy emulators for scientific simulations with deep neural architecture search},
	doi = {10.1088/2632-2153/ac3ffa},
	year = 2021,
	month = {dec},
	publisher = {{IOP} Publishing},
	volume = {3},
	number = {1},
	pages = {015013},

	journal = {Machine Learning: Science and Technology}

}

@misc{bassey2021survey,
      title={A Survey of Complex-Valued Neural Networks}, 
      author={Joshua Bassey and Lijun Qian and Xianfang Li},
      year={2021},
      eprint={2101.12249},
      archivePrefix={arXiv},
}

@article{zhang2017shaping,
	year = 2017,
	month = {jun},
	publisher = {Springer Science and Business Media {LLC}},
	volume = {7},
	number = {1},
	author = {Qian Zhang and Xiang Wan and Shuo Liu and Jia Yuan Yin and Lei Zhang and Tie Jun Cui},
	title = {Shaping electromagnetic waves using software-automatically-designed metasurfaces},
	journal = {Scientific Reports}
}

@article{Omar2021deep,
author = {Khatib, Omar and Ren, Simiao and Malof, Jordan and Padilla, Willie J.},
title = {Deep Learning the Electromagnetic Properties of Metamaterials—A Comprehensive Review},
journal = {Advanced Functional Materials},
volume = {31},
number = {31},
pages = {2101748},
doi = {https://doi.org/10.1002/adfm.202101748},
year = {2021}
}

@article{Kasim_2021,
	doi = {10.1088/2632-2153/ac3ffa},
	year = 2021,
	month = {dec},
	publisher = {{IOP} Publishing},
	volume = {3},
	number = {1},
	pages = {015013},
	author = {M F Kasim and D Watson-Parris and L Deaconu and S Oliver and P Hatfield and D H Froula and G Gregori and M Jarvis and S Khatiwala and J Korenaga and J Topp-Mugglestone and E Viezzer and S M Vinko},
	title = {Building high accuracy emulators for scientific simulations with deep neural architecture search},
	journal = {Machine Learning: Science and Technology}

}
}
\end{document}
\fi
\ifx\allfiles\undefined
\input{format/packages-IOP}
\begin{document}
\fi

\section{The SUTD-PRCM dataset}
Our SUTD-PRCM dataset was created by automating the full wave EM simulation with the SUMULIA CST Studio Suite and MATLAB to provide accurate characterization of randomly created metasurfaces.
In this section, we will introduce the dataset by providing the details of the EM simulation and the characteristics of this dataset.
\subsection{Generation of data}
An input metasurface pattern, $\mathcal{I}$, is associated with a set of densely sampled EM spectral responses under a given setting: $\mathcal{E}$, which is described by a set of physical parameters such as material type (dielectric or metal), frequency range (GHz to THz), feature size, and others.
Different settings of $\mathcal{E}$ will led to different EM responses governed by the Maxwell equations.
Therefore, a DL model can be perceived as a surrogate model that mimic the input/output behaviour of the complex systems governed by the Maxwell equations.
If a DL model is trained properly with a sufficiently large and appropriate dataset, it can be used to predict the EM response (of particular application) without solving the Maxwell equations.
Thus the quality of the dataset is the most important factor for DL models to function properly.

\begin{figure}[t]
    \centering
    \includegraphics[width=0.6\linewidth]{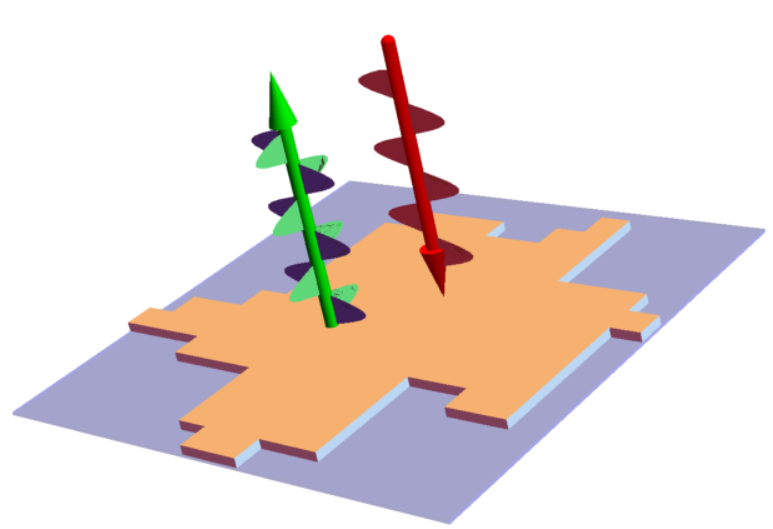}
    \caption{A $x$-polarized electromagnetic wave (red) incident on a purely reflective metasurface and the $x$- and $y$- polarized electromagnetic reflected wave (green).}
    \label{diagram_inject_transmit_reflect}
\end{figure}

For the traditional methods, with the information of $\mathcal{E}$ and the input pattern $\mathcal{I}$, one can obtain the EM response accurately and robustly via a numerical solver like rigorous coupled-wave analysis (RCWA),
finite-difference time-domain method (FDTD), or finite element method (FEM).
Here, we have adopted the FEM solver in CST Studio Suite to create the samples.
Each sample of SUTD-PRCM is made up of an input of metasurface pattern, $\mathcal{I}$, and outputs of associated $x$- and $y$- polarized reflection.
Each metasurface is represented by a unique pattern encoded by a $16\times16$ matrix that made up of 0 and 1.
The binary setting of 1 or 0 corresponds respectively to the presence or absence of a square copper patch (0.5 mm x 0.5 mm x 0.018 mm) on top of a dielectric substrate with $\epsilon_r=2.65 \times (1+0.003i)$ and $\mu_r=1$, and it is backed by a 0.18-mm-thick copper plate.
The sample has a padding of 1 mm on the sides which forms the unit cell used in the simulation.
Simulations were performed with unit cell boundary condition in $x$ and $y$ direction and open boundary condition in the $z$ direction.
An $x$-polarized plane wave is incident normally from the top of the metasurface as illustrated in Fig.\ref{diagram_inject_transmit_reflect}.
In general, the EM spectral responses include reflection and transmission of different polarizations.
For simplicity, the metasurface under consideration is of pure reflective type
(note such limitation can easily be changed in the code).
Thus the EM spectral responses only contain $x$-polarized ($\mathcal{R}$) and $y$-polarized ($\mathcal{T}$) reflection associated with a given sample.
Each spectrum is computed for frequency from 2 to 12 GHz sampled for 1001 points.
We define $\mathcal{C}_{f_i}^R$ and $\mathcal{C}_{f_i}^T$ to be the numerical values (complex numbers) related to $\mathcal{R}$ and $\mathcal{T}$, respectively, where $f_i$ denotes the discrete frequency points.
The fraction of reflected energy can be calculated from
$P=|\mathcal{C}_{f_i}^R|^2+|\mathcal{C}_{f_i}^T|^2$.
Due to conservation of energy, $P$ is capped at 1 and 
$P<$  1 implies some energy absorption into the substrate.

\ifx\allfiles\undefined
\bibliographystyle{unsrt}
\bibliography{references}
\end{document}
\fi

\ifx\allfiles\undefined
\input{format/packages-Normal}
\begin{document}
\fi

\subsection{Classes of metasurfaces}
The patterns of metasurfaces are encoded into a binary matrix of size 16 $\times$ 16.
Depending on the aggregation or configuration of the pixels (see Fig. 2 below), we divide the samples in the dataset into four classes: 
(a) Polygon-like (PLG),
(b) Polygonal ring (PLR),
(c) Pattern-combination (PTN), and
(d) Random (RDN).
The PLG class resembles filled polygon, which are connected topologically and can deform smoothly to each other. 
The PLG class is dense and are more commonly encountered in manufacturing type design \cite{haninverse,liu2018generative,sajedian2019finding}. 
The number of possible combinations of the PLG class is estimated to be about $2^{152}$.
The PLR class resembles polygonal rings, which are formed by an enclosed area in between two cocentric polygons. 
Note the inner polygon of PLR class may vanish (zero pixel), for which PLR will become PLG. 
The PTN class contains patterns formed by combining disjointedly any number of the six basic shapes such as square (9 pixels), cross (5 pixels), triangle (4 pixels) of four directions, U-shape (5 pixels), and H-shape (7 pixels). 
The possible number of PTN class is estimated to be about $2^{102}$.
Finally, the RDN class does not have any restriction, where the patterns are totally random binary images and thus has the highest number of combination at about $2^{256}$. 
Table \ref{table-classes-metasurface} summarizes the characteristics of these 4 classes of metasurfaces with a total of 260,000 samples.


\begin{figure}[ht]
    \centering
      \begin{subfigure}{.2\linewidth}
          \includegraphics[width=1\textwidth]{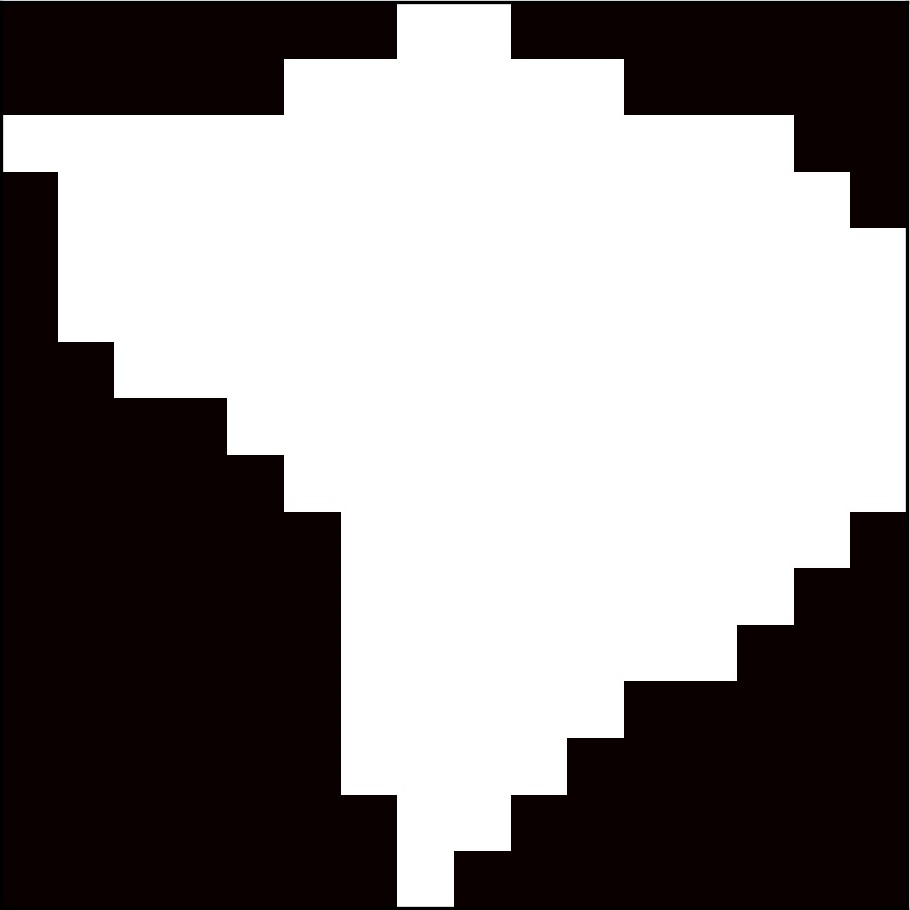}\caption{PLG}
      \end{subfigure}
      \begin{subfigure}{.2\linewidth}
          \includegraphics[width=1\textwidth]{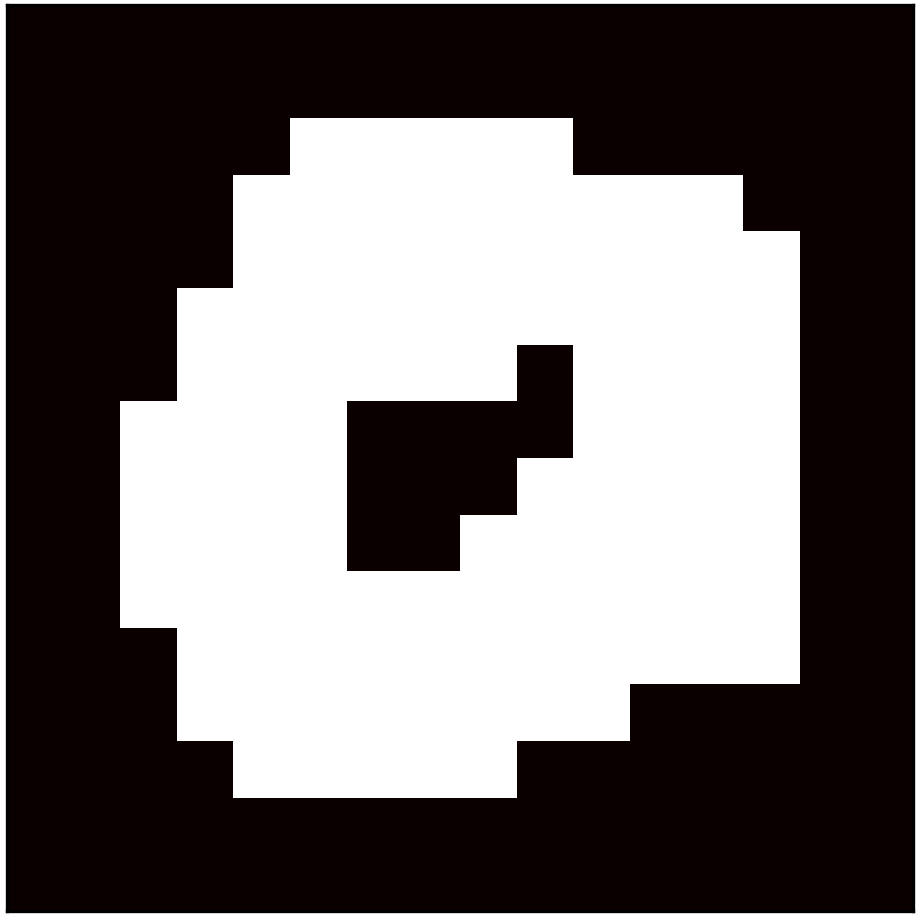}\caption{PLR}
      \end{subfigure}
      \begin{subfigure}{.2\linewidth}
          \includegraphics[width=1\textwidth]{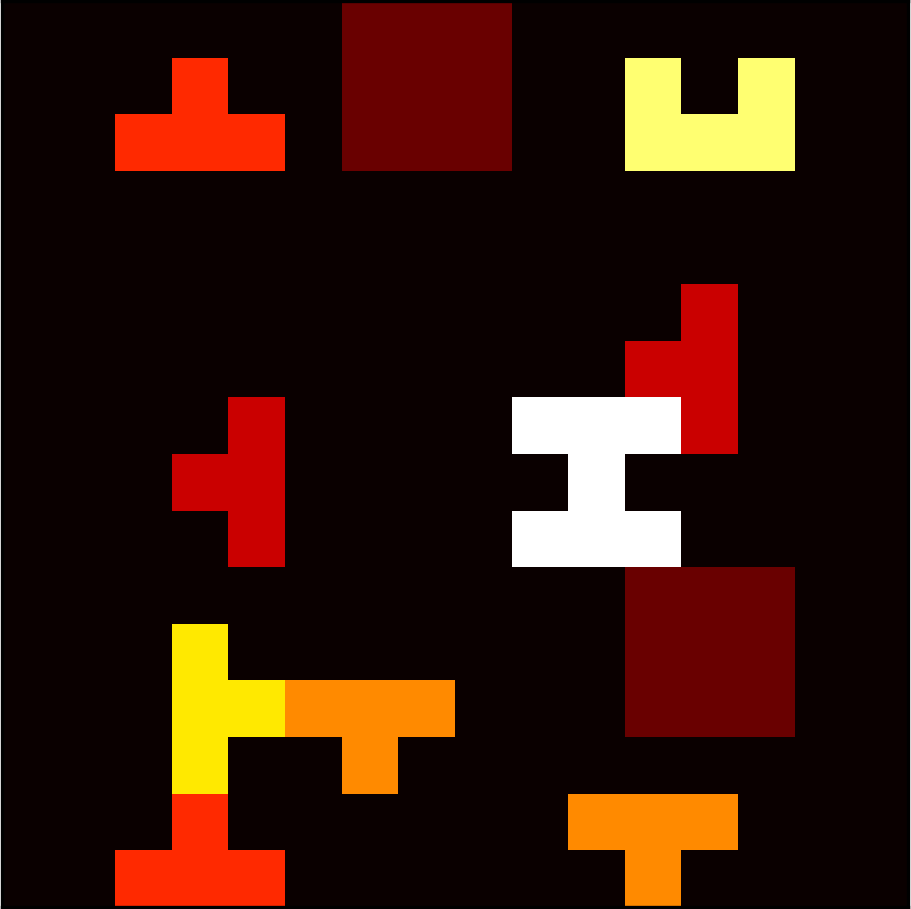}\caption{PTN}
      \end{subfigure}
      \begin{subfigure}{.2\linewidth}
          \includegraphics[width=1\textwidth]{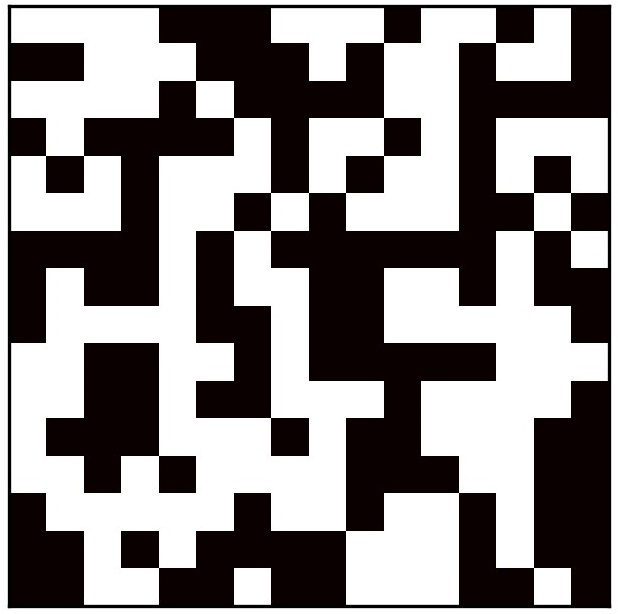}\caption{RDN}
      \end{subfigure}
      \caption{Samples of metasurface pattern from different classes.}
\end{figure}

\begin{table}[ht]
  \centering
  \caption{Classess of metasurface patterns.}
  \begin{tabular}{lrp{12cm}}
    \hline
    \textbf{Name}&        Samples    &    Description    \\ \hline
    \textbf{PLG} &  \textbf{30000}  & Polygon patterns \\
    \textbf{PLR} &  \textbf{60000}  & Polygonal ring formed by two cocentric polygons \\
    \textbf{PTN} &  \textbf{60000}  & Combination of squares, crosss, triangles, U-shapes, H-shapes \\
    \textbf{RDN} & \textbf{110000}  & Totally random binary patterns \\\hline
  \end{tabular}
  \label{table-classes-metasurface}
\end{table}

\subsection{Relationship between classes}
Within the dataset, there is no overlapping samples between any two classes.
However, the domain of the 4 classes are related.
When the inner polygon in PLR class vanishes, the patterns resembles those in PLG, thus the domain of PLG class is a subset of the PLR class.
Since the RDN class does not contain any restriction and cover all possible patterns allowed for a 16x16 binary image, the domain is a superset of the other 3 classes. 
Intuitively, we hope a DL model trained only by the RDN class of data is able to acquire sufficient information to allow equal prediction as compared to those trained separately by the other 3 classes (PLG, PLR, and PTN).
If this goal is met, we will consider that the DL model is successful, which may be able to capture the underlying physics.
Unfortunately, we have concluded in a recent study \cite{zhang2021deep} that current CNN based DL models are insufficient to reach this optimal condition for which cross classes forward prediction shows deteriorating performance.
We postulate that this is due to non-optimal neural architecture adopted, which motivates us to share our dataset in this paper with other researchers for future improvements.

\ifx\allfiles\undefined
\end{document}
\fi
\ifx\allfiles\undefined
\input{format/packages-Normal}
\begin{document}
\fi

\section{Formulations of supervised learning task}
In this section, we will demonstrate some formulations for supervised learning task based on the dataset. 
The goal is to establish a model, which is able to predict the reflection of both polarizations simultaneously. 
An obvious method is to consider the 2-branches of complex-valued output as a 4-branches of real-value output and use a gigantic model to model the data regardless of the model size, layers, and branches.
However, it is always preferable to have a more compact and efficient model that leads to less memory footprint and quicker calculation. 
To better study the ML algorithm's behavior on this dataset, we present some simpler EM characteristics, which can be derived from the dataset.

\subsection{Complex response}
Instead of aiming to predict the reflection of both polarizations, divide and conquer can be a good strategy to get started. 
We focus on predicting just one polarization at a time, either the $x$- or $y$-polarized reflection. 
In doing so, the complex number based EM response still remains unconventional as compared to the traditional DL methods in dealing with real-number based datasets.
Without a viable complex-valued neural network available to apply directly, an improvised solution is often to drop the phase information or to split the real and imaginary part into two separate components during the training process.
Note that this is equivalent to ignoring the inherent relationship between the real and imaginary parts.
This might not be the ideal approach (it is inconsistent in terms of physics) but we considered it is an adaptive measure to quickly tap into the existing DL models that we can use directly.
We note that there is a growing interest in exploring the advantages of complex-valued neural networks \cite{bassey2021survey}, which may be used for our complex-number based SUTD-PRCM datasets shared here, and this will be studied in a future work.

\subsection{Magnitude and phase spectrum}
\begin{figure}[ht]
  \centering
  \begin{subfigure}{.35\textheight}
      \includegraphics[width=1\textwidth]{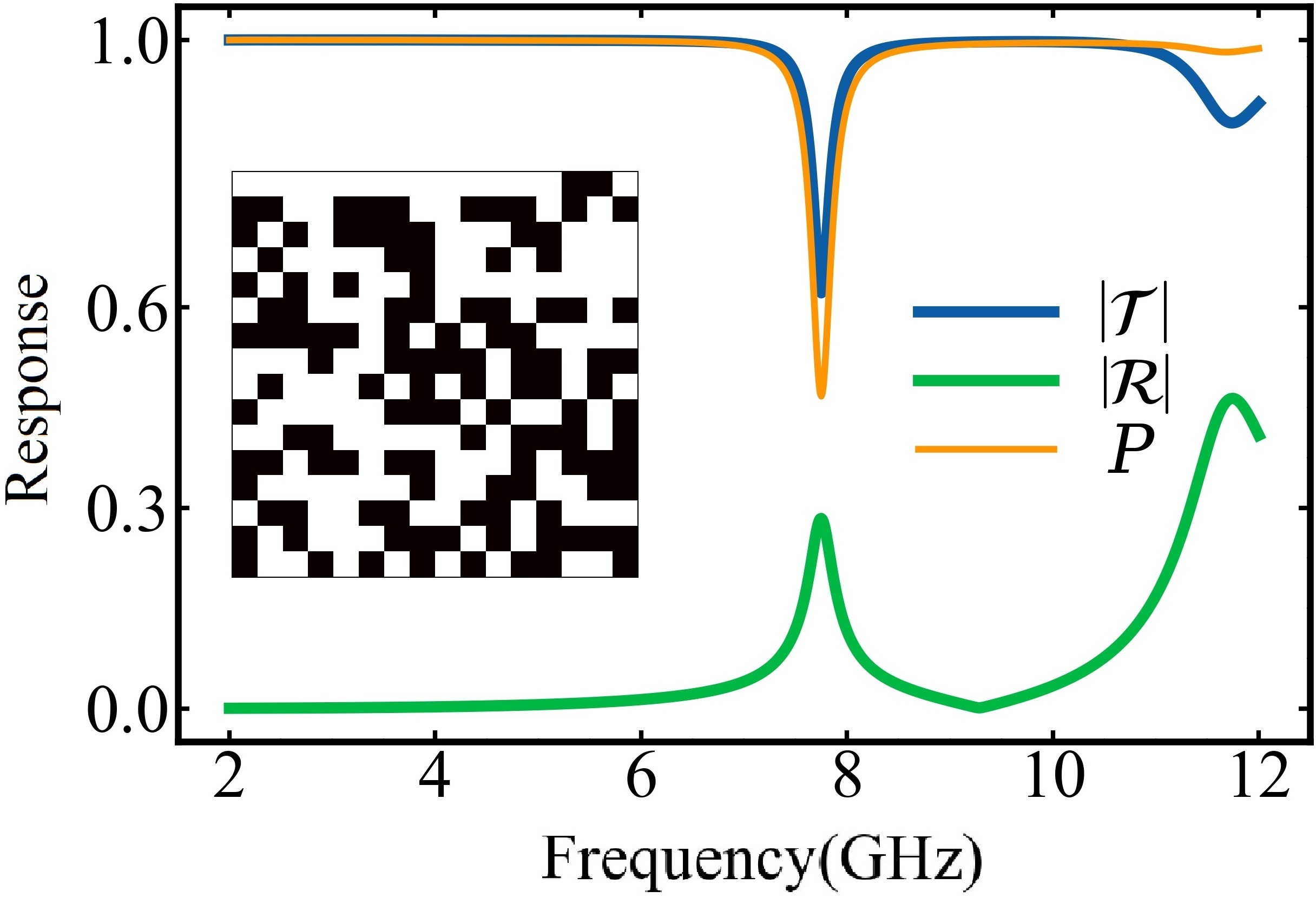}\caption{}
      \label{magnitudedataset} 
  \end{subfigure}
  \begin{subfigure}{.3\textheight}
  \includegraphics[width=1\textwidth]{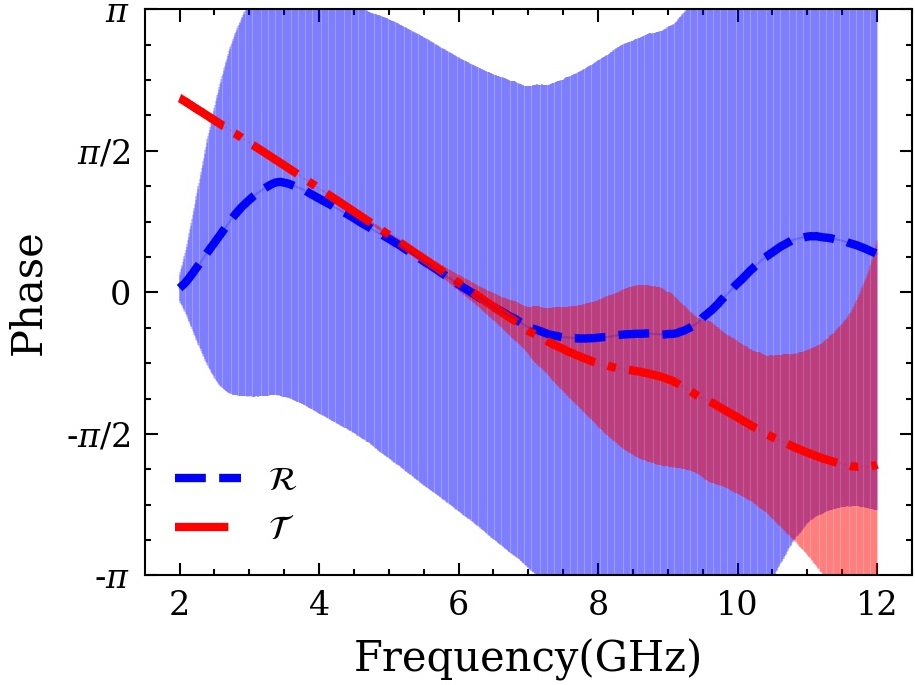}\caption{}\label{phase-curve-mean-var}
  \end{subfigure}
  \caption{(a) Magnitude spectra of a RDN metasurface sample: $|\mathcal{T}|$, $|\mathcal{R}|$, and $P$ = $|\mathcal{T}|^2$ + $|\mathcal{R}|^2$. (b) The mean and variance of phase spectrum for the entire dataset.} 
\end{figure}

For simplicity, the complex-valued EM response from SUTD-PRCM dataset is converted into two real-number based representation of magnitude (Fig.\ref{magnitudedataset}) and phase (Fig.\ref{phase-curve-mean-var}).
In Fig. \ref{magnitudedataset}, for a given RDN sammple, we show the magnitude of $x$-polarized reflection ($|\mathcal{R}|$), $y$-polarized reflection ($|\mathcal{T}|$), and $P$ = $|\mathcal{T}|^2$ + $|\mathcal{R}|^2$ for a spectrum from 2 to 12 GHz.
In Fig. \ref{phase-curve-mean-var}, we show the statistical plotting of the mean and variance of the phase for $\mathcal{R}$ ($x$-polarization in blue) and $\mathcal{T}$ ($y$-polarization in red).
%
%
Within this dataset, the magnitude spectra are typically continuous and smooth.
Compressed representations for these spectra are highly desirable in reducing the number of parameters, which will improve the efficiency of training and accuracy of model prediction. 
Fourier transform\cite{lin2020inverse,bruna2013invariant}, discrete cosine transform, wavelet transform and uniform down-sampling are some popular options in constructing compressed representations. 
In our prior work \cite{zhang2021deep}, we reported that uniform down-sampling has produced good results to predict the magnitude.
However, the phase spectra are oscillatory due to the periodicity of $2\pi$.
It is evident that the variance is high for the $x$-polarized reflection, $\mathcal{R}$. 
This behaviour can be problematic in training a DL model and leads to biased preference of real/imaginary representation in some studies.

\subsection{Peak locations in the spectrum}
In some applications such as filtering \cite{haninverse}, the locations of peaks in a spectrum are important.
One way to extract the locations of peaks is to convert it to the sum of the Pearson Type VII function \cite{gupta1998peak}, which will reduce the dimension of the output EM response from $1001$ to $4N+2$, where $N$ is the number of peaks.
However, it may not retain the locality information and unable to recover exactly the spectrum after the transformation.
Note that the magnitude spectra for PLG, PTN, and PLR classes are usually smooth, so these three classes are not suitable for experimentation of this approach. 
The magnitude spectra for RDN class of metasurfaces contains more peaks, which is used for this testing.
From the RDN class, it is found that there is at most six peaks ($N$ = 6) between 2 to 12 GHz, and thus the output of the model can be greatly compressed to only 26 (compared to 1001).
The general fitting function used is
$$
C(x) =o+k*x+ \sum^{Peaks}_i a_i[1+\frac{(x-d_i)^2}{b_i^2}]^{-m_i}
$$
where $d_i \in [2,12]$, $a_i \in [0,1]$, $b_i\in[0,5]$, $m_i\in[1,6]$.

\begin{figure}[ht]
  \centering
  \begin{subfigure}{.61\linewidth}
      \includegraphics[width=\textwidth]{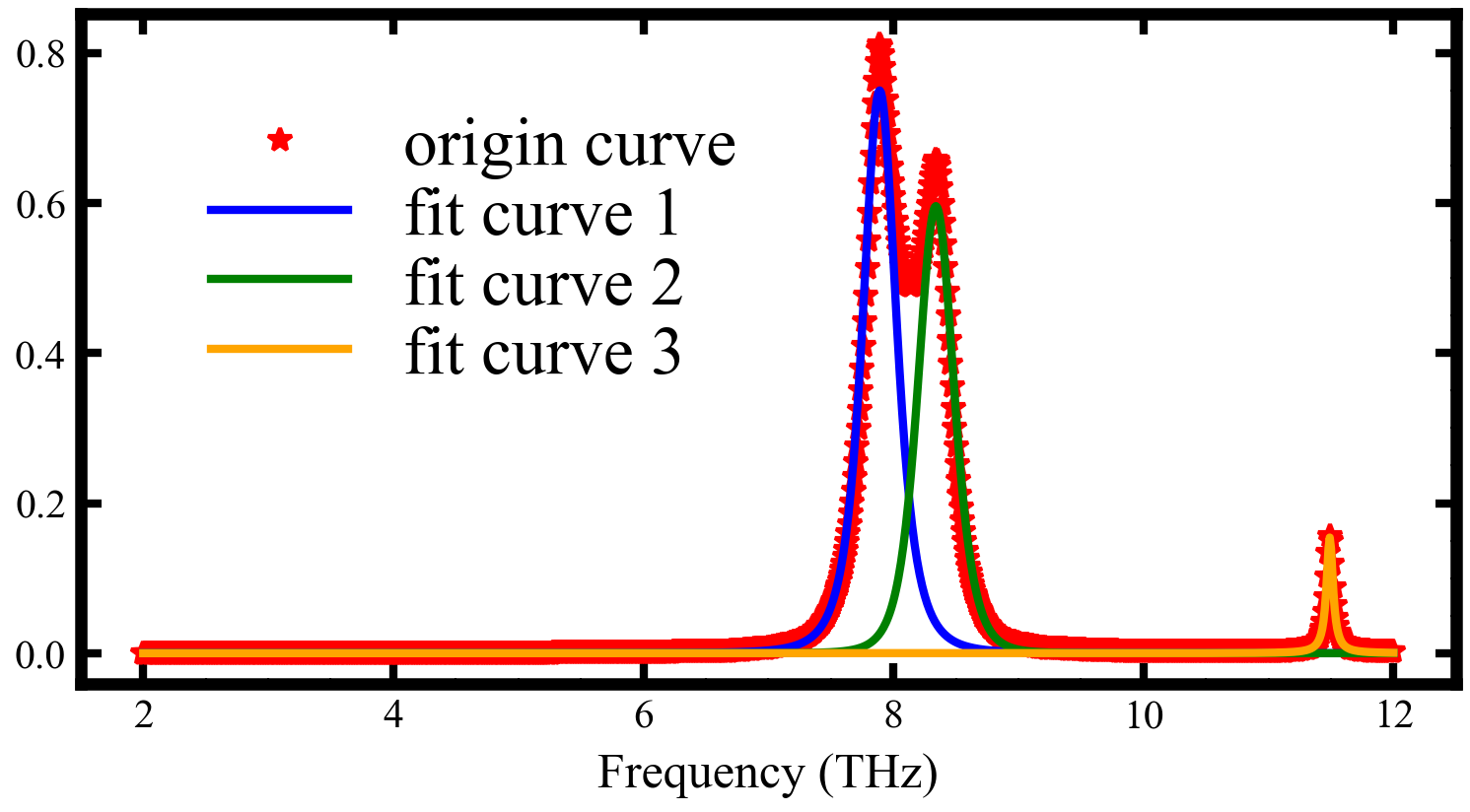}
      \caption{}
      \label{demo_for_peak_fit_curve}
  \end{subfigure}
  \begin{subfigure}{.3\linewidth}
      \includegraphics[width=\textwidth]{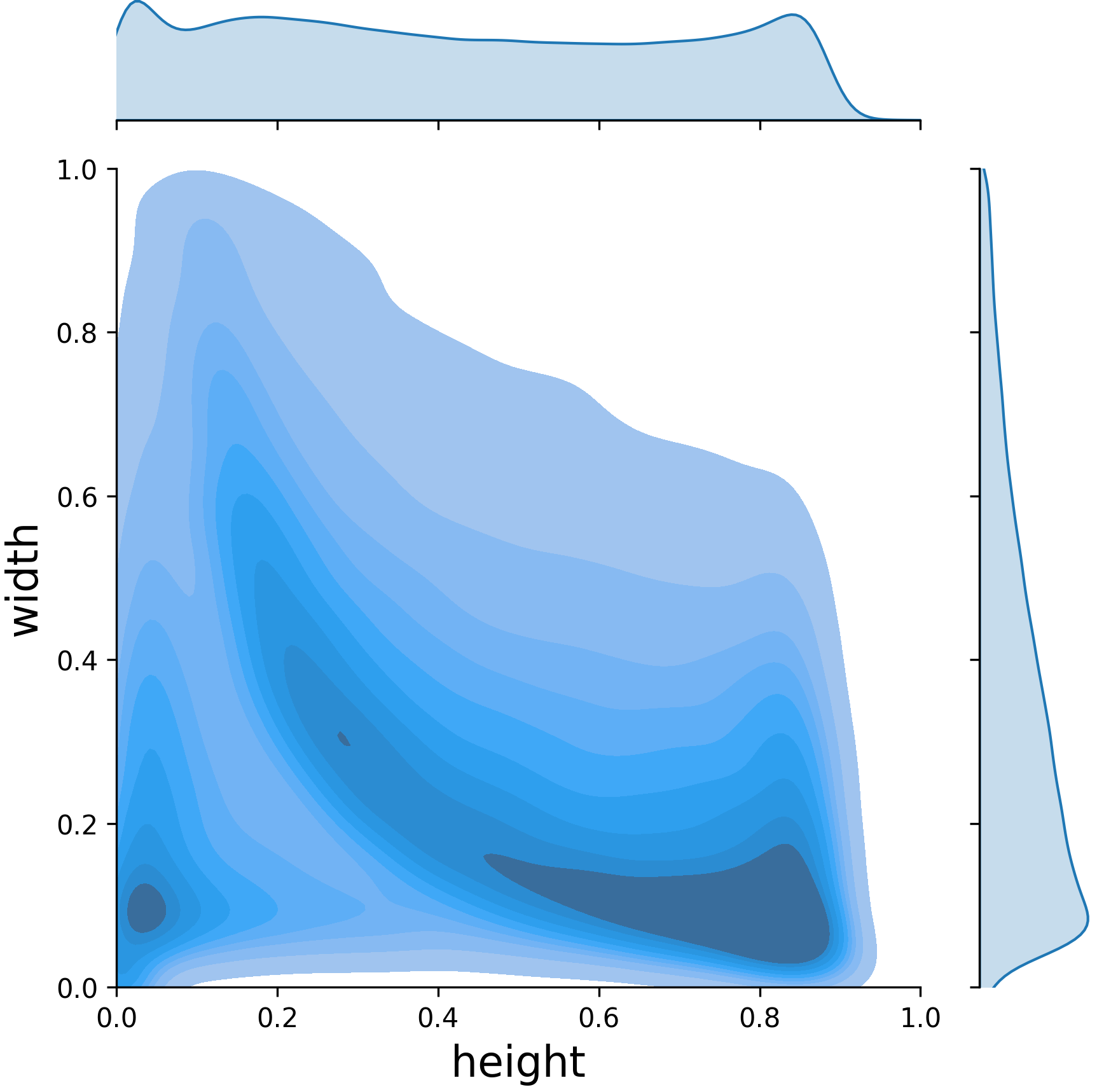}
      \caption{}
      \label{RDN_Peak_Height_to_Width}
  \end{subfigure}
  \caption{(a) A demonstration of peak decomposition performed on a magnitude spectrum using the Pearson Type VII function and the resulting 3 peaks. 
  (b) The statistical distribution of heights ($a_i$) and widths ($b_i$) of the peaks extracted from the magnitude spectra, $|\mathcal{R}|$, for all metasurfaces in the RDN class. 
  With the hotspot concentrating at the bottom right region, it is observed that the peaks are mostly distinctive sharp peaks.
  The peak decomposition is an effective method for extracting peak characteristics of the spectra. 
  } 
\end{figure}

Figure \ref{demo_for_peak_fit_curve} illustrates an example of the decomposition of an arbitrary magnitude spectrum, $|\mathcal{R}|$ with three peaks.
The parameters, $a_i$ and $b_i$, measure respectively the heights and widths of a peak. 
Fig. \ref{RDN_Peak_Height_to_Width} shows the distributions of height ($a_i$) and width ($b_i$) of all the samples in the RDN class of our dataset.
This proposed mathematical transformation has the advantage that the fitting parameters, $a_i$, $b_i$, and $m_i$ are more robust against noise, which do 
not require high precision.  
In contrast, the location parameter ($d_i$) is critical.
In order to obtain the accurate peak location, it is desirable to formulate a classification problem as compared to a regression problem.
In this case, a metasurface can be associated with multiple labels. 
Each label, $i$, belongs to integer values between 1 to 1001.
To identify the peaks, $i$ will be used to index into the discrete frequencies between 2 to 12 GHz.
In general, multi-label classification problem can be challenging.
For simplicity, we consider a binary classification problem, which we term as maximum peak binary classification (MPBC) task. 
In this learning task, we determine whether the location of the maximum peak is located at a position larger than a given threshold of frequency.
If this occurs, the associated input metasurface is assigned a positive label.
Otherwise, the input metasurface is assigned a negative label.
To avoid bias in MPBC, it is desirable to have balanced classes, i.e. both positive and negative classes possess approximately equal number of training data.
Successfully dealing with this task will help to extend the problem to a multi-label classification setting.
Alternatively, by design, multi-label classification can be realized by cascading multiple binary classifiers.
In the following section, we will demonstrate the application of NAS using this MPBC problem.

\ifx\allfiles\undefined
\bibliographystyle{unsrt}
\bibliography{references}
\end{document}
\fi
\ifx\allfiles\undefined
\input{format/packages-Normal}
\begin{document}
\fi

\section{NAS for high performance neural architecture}
In this section, we will consider the MPBC problem on RDN class mentioned above.
For benchmarking purpose, we adopt some off-the-shelf models including traditional ML models and popular DL models in CV community, and NAS will be implemented to compare with these models.
\ifx\allfiles\undefined
\end{document}
\fi
\ifx\allfiles\undefined
\input{format/packages-Normal}
\begin{document}
\fi

\subsection{Benchmark with existing ML models}
\begin{figure}[b]
  \centering
  \includegraphics[width=0.5\textwidth]{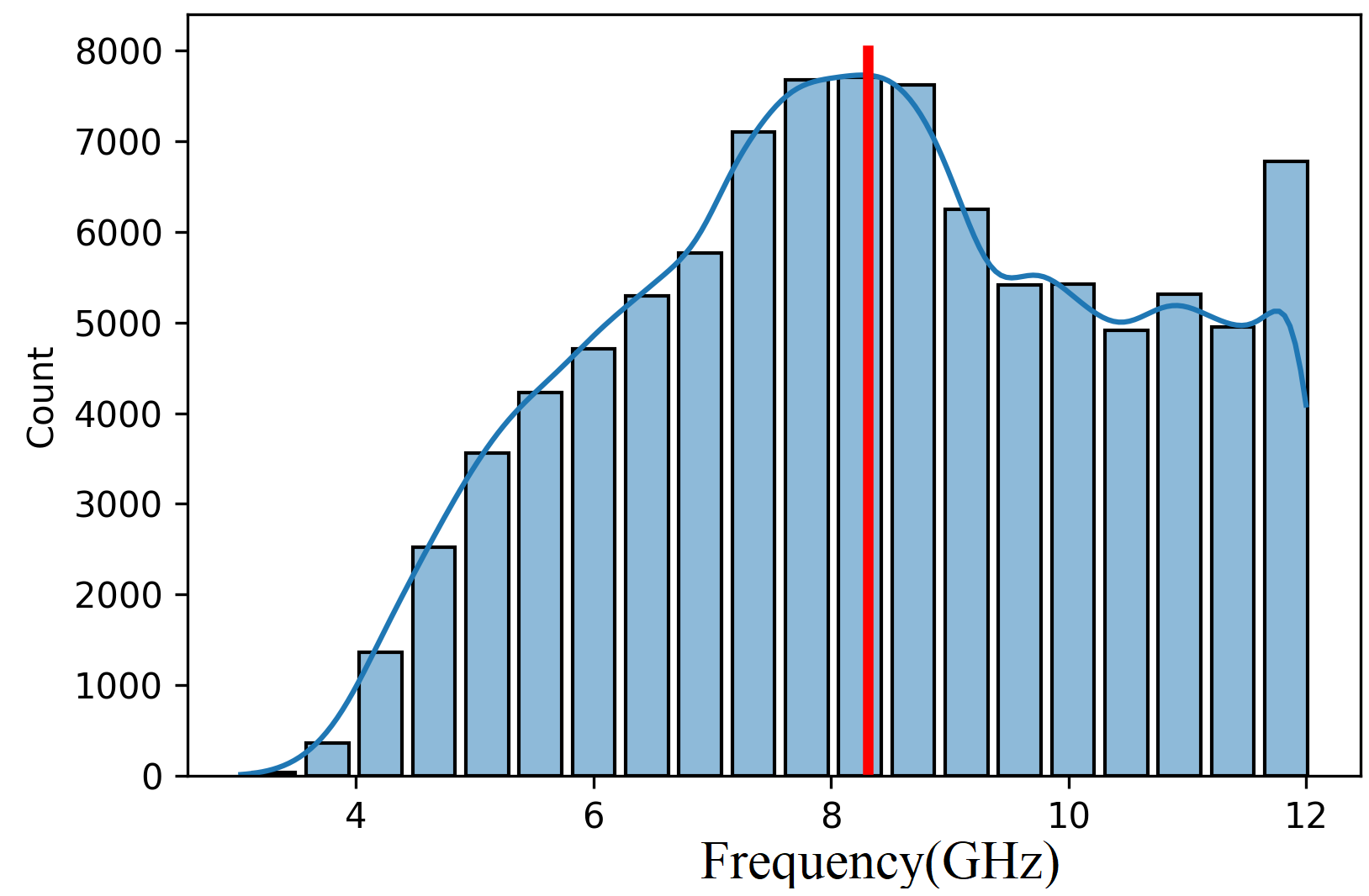}
  \caption{The histogram of location of maximum peak in the magnitude spectra, $|\mathcal{R}|$, of RDN class of metasurface.}
  \label{RDN_Max_Peak_location_hist}
\end{figure}

The benchmarking task here is to predict whether the maximum peak in a spectrum is located at a frequency larger than a threshold frequency.
If the maximum peak is at frequency larger than the threshold, the binary image is assigned a positive label, otherwise, a negative label.
The threshold frequency is selected to be 8.3 GHz, which is the median frequency as shown in the histogram in Fig. \ref{RDN_Max_Peak_location_hist}.
Focusing on the RDN class of SUTD-PRCM dataset and extracting the peak information for MPBC, a label either positive or negative can then be assigned to a metasurface pattern. 
RDN class has 110,000 samples, where 108,000 are used as training samples, and 2000 are as test samples.
In the training dataset, there are 55,364 positive labels and 52,636 negative labels.
In the test dataset, there are 1508 positive labels and 1492 negative labels.
The classification accuracy is a popular metric to assess the performance of the classification models.
A simple baseline classifier which always predicts the same label can produce a baseline accuracy of $1508/3000 = 50.27\%$ due to roughly equal positive and negative classes in the training dataset.

\begin{figure}[t]
    \centering
    \includegraphics[width=\textwidth]{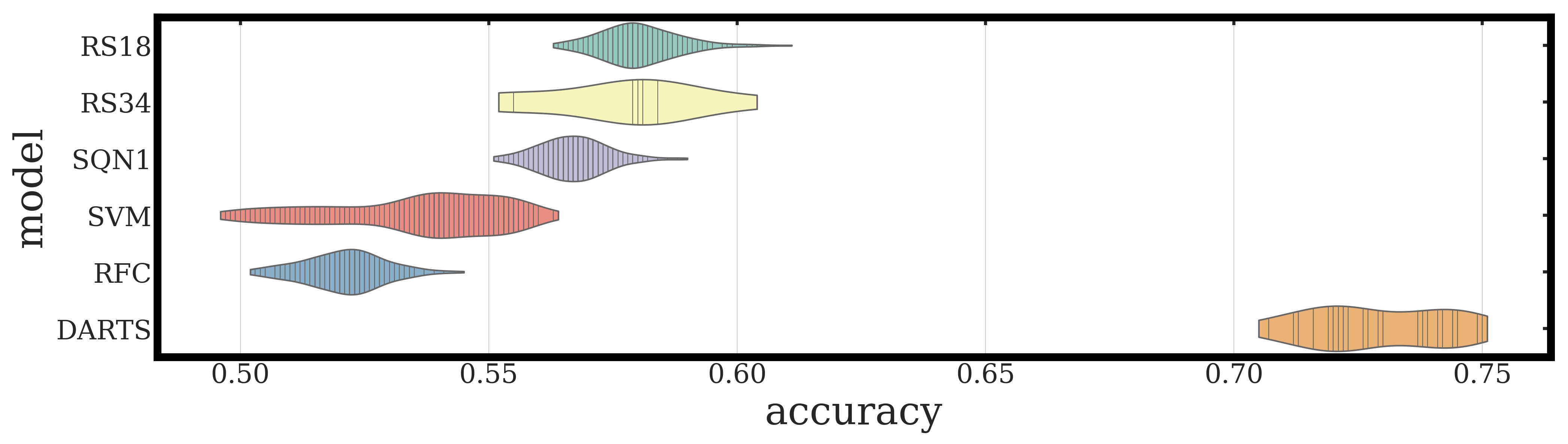}
    \caption{Every scattering point is representing one trial with different hyper parameters.
             Different colors represent different machine learning models. Each model is trained to converge. The five models (on the left) are the traditional machine learning models listed from top to bottom: RS18 (Resnet18s), RS34 (Resnet34S), SQN1 (SqueezeNet1S), SVM (Support Vector Machine), RFC (Random Forest Classification machine).
             The last one on the right is the best NAS based model (DARTS), which holds the state-of-the-art performance in MPBC task. 
        }\label{Model_banchmerk_for_RDN_Balanced_2_class_dataset}
\end{figure}

In comparison, we have applied traditional machine learning models such as random forest classifier (RFC) and linear/log support vector machine (SVM) classifier.
However, the results are only marginally better than the baseline accuracy, i.e. RFC and SVM is about 53\% and 55\%, respectively.
The location of maximum peak cannot be well identified with these ML models.  
Considering our RDN class of datasets resembles images, we apply some off-the-shelf neural network architectures, such as deep multi layer perception (MLP), Resnet18 (RS18), Resnet34 (RS34), and SqueezeNet (SQN1) to tackle this problem. 
These architectures are well known to perform excellent with CV related tasks, and  they have been adapted to accommodate the image size in the dataset and trained from scratch.
To our surprise, all these neural network models do not score above 60$\%$ accuracy as shown in Fig.\ref{Model_banchmerk_for_RDN_Balanced_2_class_dataset}.
Notice that another model labelled DARTS is included in the same figure, which performs significantly better.
This model is based on neural architecture search (NAS) that we will elaborate in the next section.
Note in our prior work \cite{zhang2021deep}, a modified version of Resnet18 was reported to achieve excellent performance in a related regression problem formulated to predict EM response of the RDN class in SUTD-PRCM dataset.

The low accuracy (less than $60\%$) certainly is not appealing and definitely signifies more research to be done.
For comparison, we applied the same set of ML/DL models to a scaled binarized MNIST dataset, which is a well studied handwritten digit dataset.
Traditional ML models can achieve over 90\% accuracy: SVM (91\%) and RFC (96.9\%).
DL models can achieve over 95\% accuracy: SqueezeNet1S (95.5\%), Resnet18S (96\%), DARTS (97\%), and Resnet34S (98.5\%).
As of writing, the best DL model on the standard MNIST dataset is Efficient-CapsNet \cite{mazzia2021efficient}, which can attain a classification accuracy of 99.9\%. 
Such a good performance is credited to decades of effort from ML community in testing on the same common dataset, which reinforces the importance of having a common dataset for metasurfaces if an optimal DL model is aimed to be developed for the design of complex metasurfaces.
Thus, this motivates us to share our dataset (SUTD-PRCM) in this paper.


While the classification problem using RDN class here might appear to be similar as compared to MNIST dataset, and one may wonder if an accuracy of over 90$\%$ is possible using the known DL models. 
However, we emphasize here the significant differences between random metasurfaces (RDN) and handwritten digits (MNIST).  
Each image of MNIST is a centralized and continuous image. 
Furthermore, each digit (0-9) can smoothly deform to each other.
This restricts the possible patterns to a small subset of the domain of a $16\times 16$ binary image. 
Contrary to the MPBC problem, the input pattern can be any random binary $16\times 16$ image. 
Apparently, the complexity of MPBC problem studied here in using our dataset is higher than that of the MNIST digit recognition.
With the initial findings from \cite{zhang2021deep}, we speculate that the architecture of the existing neural network used for SUTD-PRCM dataset might not be optimal yet.
In the following section, we demonstrate further improvements in this direction.

\ifx\allfiles\undefined
\end{document}
\fi
\ifx\allfiles\undefined
\input{format/packages-Normal}
\begin{document}
\fi
\subsection{Neural architecure search}
    Neural architecture search (NAS) is certainly not a new concept but it has just become an affordable tool in everyone's ML toolbox.
    In fact, NAS is a follow-up idea to automate the laborious effort to design optimal neural architecture for any given problem in DL.
    Early attempts has resorted to huge amount of computational resources that only big enterprises or research groups can afford.
    It is not until recently that one can perform NAS on a common workstation \cite{qin2019nasnet,zoph2016neural,liu2018darts,kandasamy2018neural}.
    
\begin{figure}[ht]
    \centering
    \begin{subfigure}{0.95\linewidth}
        \includegraphics[width=1\textwidth]{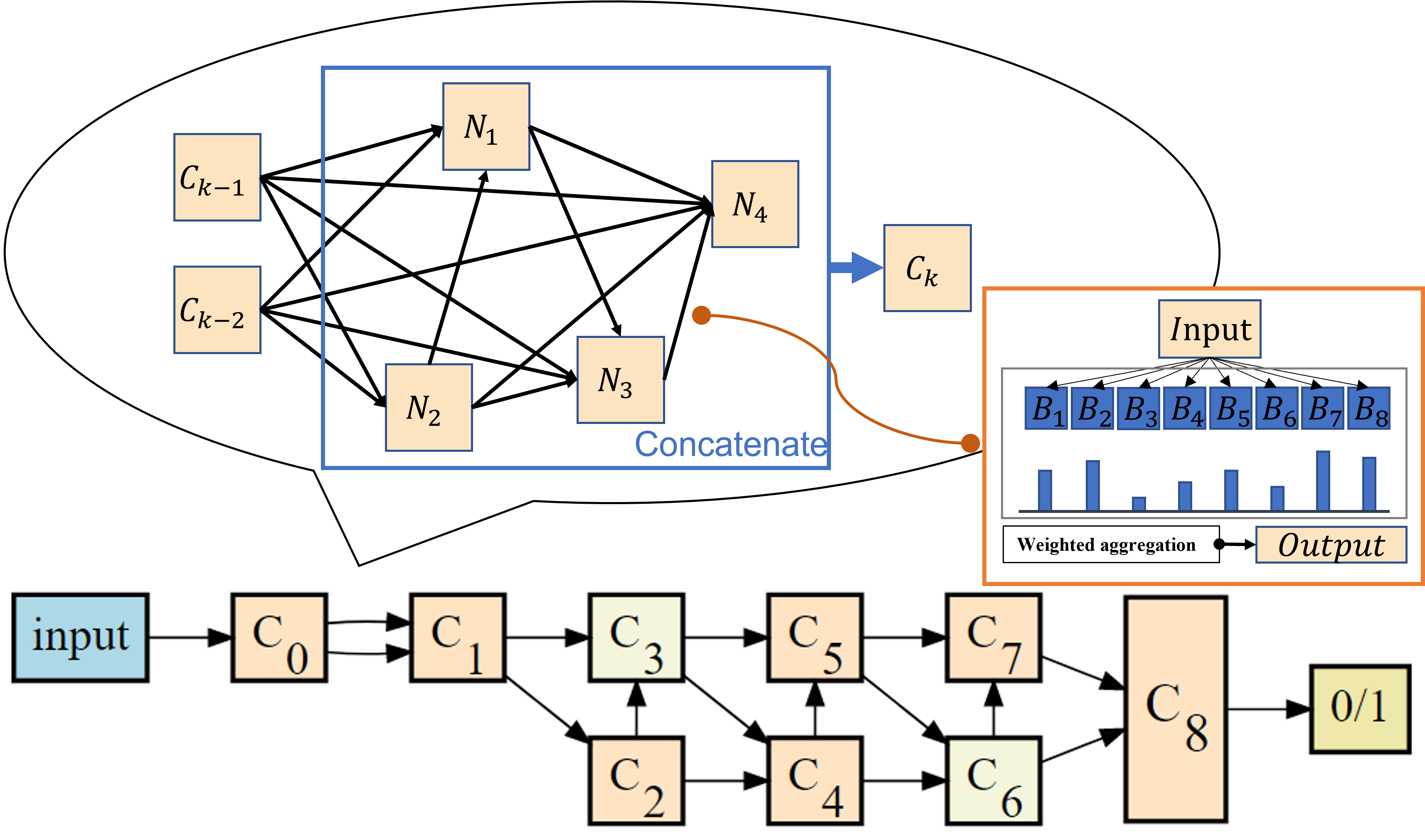}
        \caption{Main structure containing normal (orange) and reduce (yellow) branch.}
        \label{DARTS_model_full}
    \end{subfigure}
    \begin{subfigure}{1\linewidth}
        \includegraphics[width=1\textwidth]{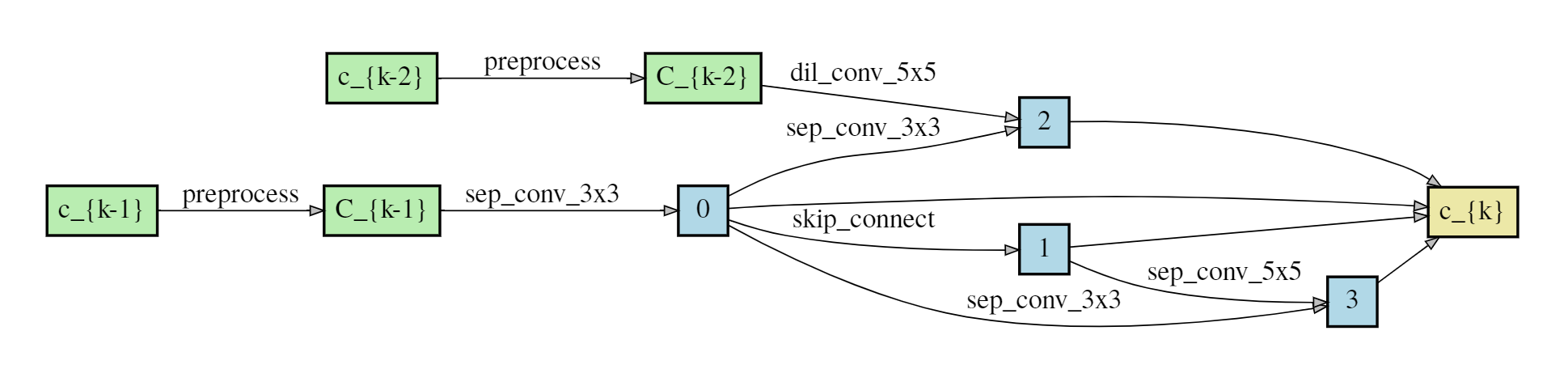}
        \caption{Normal branch.}
        \label{PC_DARTS_metas_d_normal}
    \end{subfigure}
    \begin{subfigure}{1\linewidth}
        \includegraphics[width=1\textwidth]{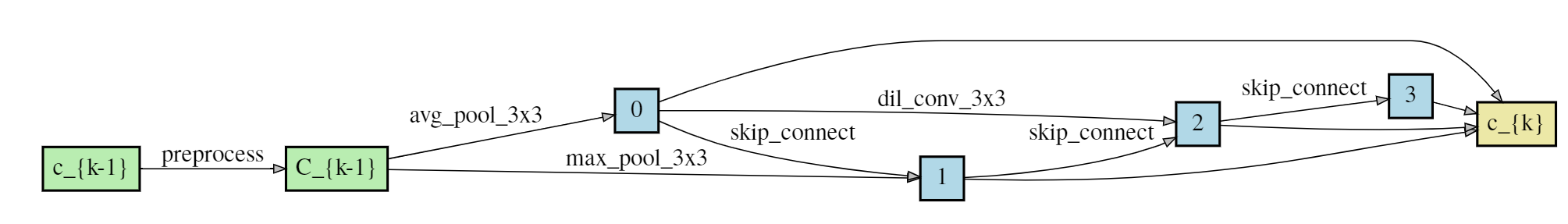}
        \caption{Reduce branch.}
        \label{PC_DARTS_metas_d_reduction}
    \end{subfigure}
    \caption{(a) The main structure of the network architecture search (NAS) module. The orange and yellow square represent the two architectural types respectively. There are nine cells from $C_0$ to $C_8$. Each cell is the interaction of the last two cells. The interaction contains 4 intermediate units ($N_1$ to $N4$). Each unit is the sum aggregation of several mapping (black arrow). There are 14 mappings in one interaction. Each mapping is a weighted aggregation of candidate operations. In this paper, we use eight symmetry operations $S_i$. A final linear layer is applied to convert $C_8$ to the binary output of positive $1$ or negative $0$. (b) and (c) are the final truncated architecture produced by NAS. Only the first and the second important path are kept for each node. 
    }\label{DARTS_model}
    \end{figure}

    The objective of NAS is to discover the best architecture for a neural network tailored for a specific requirement based on a given dataset. 
    It essentially takes the process of a human manual process in modifying neural network for better performance. 
    Thus NAS is an automated discovery of more optimal network architectures for a given dataset. 
    It represents a set of tools and methods that will test and evaluate a large number of architectures across a search space using a search strategy in order to select the one that is most suitable for a given problem by maximizing a fitness function.
    The most well-known NAS method is the Google's NASNet \cite{qin2019nasnet}, but this method requires thousands of TPU/GPU resources that are not affordable for common research groups.
    In the rapid advancement of NAS, researchers have put forward many experimental NAS methods like Reinforcement Learning (RL) Methods \cite{zoph2016neural}, Gradient-based (GB) Methods \cite{liu2018darts}, Evolutionary Algorithms (EA) \cite{real2017large} and Bayesian Optimization (BO) \cite{kandasamy2018neural}.
    In this section, we adopt the Differentiable Architecture Search (DARTS) with Geometry-Awared gradient algorithm \cite{li2020geometry,liu2018darts,xu2019pc} as it requires significantly less computational resources as compared to other NAS methods.
    Unlike the RL or EA approaches, DARTS introduces a continuous relaxation scheme that enables differentiable learning objective.
    This differentiability is the key to the computational feasibility following a gradient based approach.
     In our experiment, we inherit the spirit of \cite{liu2018darts} and update the architecture following a geometry strategy \cite{li2020geometry}, which helps us to converge quickly and escape from the local minima. The full search space is shown in Fig. \ref{DARTS_model_full}.
    It is noted that we modify some configurations to adapt to the MPBC problem.
    The candidate architecture finally converges into a relatively simple structure shown in Fig. \ref{PC_DARTS_metas_d_normal} and Fig. \ref{PC_DARTS_metas_d_reduction}.
    


    Our finding shows the NAS based DARTS model is able to achieve an accuracy over $75\%$ as shown in Fig.\ref{Model_banchmerk_for_RDN_Balanced_2_class_dataset}, which surpasses all the ML models in our experiment.
    This optimal network architecture suggested by NAS-DARTS requires $233,682$ parameters, which is only about $1/5$ and $1/10$ of the requirements from Resnet18S and Resnet34S, respectively [see Table \ref{parameter_comparision}].
    Intuitively, more parameters that accompanying by a bigger model provide better capability to capture the inherent relationship between inputs and outputs.
    Surprisingly, the NAS-DARTS approach suggests an alternative shallow and smaller architecture to achieve better performance than the other models.
    Table \ref{parameter_comparision} shows a comparison of the number of parameters and operations of different models used in the SUTD-PRCM (RDN Class) dataset.
    \begin{table}[t]
        \centering
        \caption{The comparison between different neural architectures and NAS based neural architecture.
        The parameters are referring to the total free parameters in all operations.
        For example, convolution with kernel size $(C_2,C_1,w,h)$ is one operation with $w\times h \times C_1 \times C_2$ parameters.}
        \begin{tabular}{|l|c|r|}
            \hline
            Model                   & No. Operations    & No. Parameters       \\ \hline
            {\textbf{SqueezeNet1S}} & { 51}  & { 742306}   \\
            {\textbf{MLP}}          & { 30}  & { 5659074}  \\
            {\textbf{Resnet18S}}    & { 62}  & { 11174338} \\
            {\textbf{Resnet34S}}    & { 110} & { 21282498} \\\hline
            {\textbf{DARTS}}       & { 194} & { 232034}   \\
						\hline
        \end{tabular}

        \label{parameter_comparision}
    \end{table}

    The trainable parameters of the DARTS model are much less than the traditional DL model, but it has double or triple the operation numbers, which highlights the importance of a suitable neural network architecture.
    Designing a suitable meta-operation may be more effective than building an arbitrary large and deep neural network architecture.
    The detailed structure of our NAS-based DL model (Fig.\ref{DARTS_model}) has shown that convolution stacking is not the dominant element of the architecture anymore, which implies that low-level features are preferred over deep hierarchical high-level features (that are common for traditional CNN), thus explains why the CNN based models like Resnet and SqueezeNet are not performing well in this SUTD-PRCM dataset studied in this paper.

    \ifx\allfiles\undefined
\end{document}
\fi
\ifx\allfiles\undefined
\input{packages}
\begin{document}
\fi

\section{Conclusion}
We have presented our home-made SUTD-PRCM dataset based on numerical EM simulations of metasurfaces that contains EM spectral responses associated with complex metasurface patterns, which are divided into four classes having different properties.
This dataset has been tested in a prior work \cite{zhang2021deep} and we postulate that existing CNN based DL models is likely not the optimal architecture yet.
Thus we are sharing this dataset with the community for further improvement.
This SUTD-PRCM may also be used as a common dataset of random metasurfaces for more quantitative performance comparison between different models and training strategies. 

In the second part of the paper, we discuss a binary classification problem formulated based on this dataset. 
By using the Differentiable Architecture Search (DARTS) based neural architecure search (NAS) method, we show the improvement over the traditional ML methods and off-the-shelf DL models (popular in the computer vision community) such as Resnet and SqueezeNet.
Based on NAS-DARTS results, this SUTD-PRCM dataset prefer shallow and wide neural networks for better prediction.
With modern approach of inverse design often features a fast surrogate model in terms of DNN, this finding has profound impact on the design applications of complex metasurfaces as well.
In future work, one can probably improve by fine-tunning the NAS method, which is beyond the scope of this paper.
In particular, making tradeoff for search space of neural architecture can be tedious and frustrating. 
Through performing exploratory analysis with this physics-based SUTD-PRCM dataset, we also hope to alleviate some of the difficulties and this will be investigated in our future works, including in using complex number based DL models.
\ifx\allfiles\undefined
\end{document}
\fi

\section*{Data availability statement}
The data that support the findings of this study are openly available at the following URL/DOI: https://github.com/veya2ztn/SUTD\_PRCM\_dataset

\ack{This work was supported by USA Office of Naval Research Global (N62909-19-1-2047) and SUTD-ZJU Visiting Professor (VP 201303).
T.Z acknowledges the support of Singapore Ministry of Education PhD Research Scholarship.
Y.S.A. acknowledges the support of SUTD Start-Up Research Grant (SRT3CI21163). }

\clearpage

\bibliographystyle{unsrt}


\newcommand\newblock{\hskip .11em\@plus.33em\@minus.07em}

\end{document}